\documentclass[journal]{IEEEtran}

\usepackage{orcidlink}
\usepackage{csquotes}
\usepackage{graphicx}
\usepackage{soul}
\usepackage{float}
\usepackage{multirow}
\usepackage{booktabs}
\DeclareUnicodeCharacter{0301}{\'{e}}
\usepackage{amsmath}
\usepackage{amssymb}
\usepackage[normalem]{ulem}
\usepackage{xspace}
\usepackage{makecell}
\usepackage[table]{xcolor}
\definecolor{Gray}{gray}{0.85}

\begin{document}

\title{From Satellite to Street: A Hybrid Framework Integrating Stable Diffusion and PanoGAN for Consistent Cross-View Synthesis}

\author{%
    Khawlah Bajbaa, 
    \thanks{Khawlah Bajbaa is with the Department of Information and Computer Science, King Fahd University of Petroleum and Minerals, Dhahran, 31261, Kingdom of Saudi Arabia (email: g202115030@kfupm.edu.sa).}
    Abbas Anwar, 
    Muhammad Saqib\,\orcidlink{0000-0003-4374-0888},  
    \thanks{Muhammad Saqib is with NCMI, CSIRO, Marsfield, Sydney, Australia (corresponding author, email: muhammad.saqib@csiro.au).}
    Hafeez Anwar\,\orcidlink{0000-0001-9529-3966},
    \thanks{Abbas Anwar and Hafeez Anwar are with the Department of Computer Science, National University of Computer and Emerging Sciences (FAST-NUCES), Peshawar, 25000, Pakistan (email: hafeez.anwar@nu.edu.pk).} 
    Nabin Sharma\ \orcidlink{0000-0003-0841-1245}
    \thanks{Nabin Sharma is with Faculty of Engineering and IT, University of Technology Sydney, Australia (email: nabin.sharma@uts.edu.au).}
    and Muhammad Usman\ \orcidlink{0000-0003-2059-7206}
    \thanks{Muhammad Usman is with Faculty of Science, University of Ontario Institute of Technology, Canada (email: muhammad.usman8@ontariotechu.ca).}
 
}

\markboth{}%
{Shell \MakeLowercase{\textit{et al.}}: A Sample Article Using IEEEtran.cls for IEEE Journals}

\maketitle
\begin{abstract}
Street view imagery has become an essential source for geospatial data collection and urban analytics, enabling the extraction of valuable insights that support informed decision-making. However, synthesizing street-view images from corresponding satellite imagery presents significant challenges due to substantial differences in appearance and viewing perspective between these two domains. 
This paper presents a hybrid framework that integrates diffusion-based models and conditional generative adversarial networks to generate geographically consistent street-view images from satellite imagery.
Our approach uses a multi-stage training strategy that incorporates Stable Diffusion as the core component within a dual-branch architecture. To enhance the framework's capabilities, we integrate a conditional Generative Adversarial Network (GAN) that enables the generation of geographically consistent panoramic street views. Furthermore, we implement a fusion strategy that leverages the strengths of both models to create robust representations, thereby improving the geometric consistency and visual quality of the generated street-view images.
The proposed framework is evaluated on the challenging Cross-View USA (CVUSA) dataset, a standard benchmark for cross-view image synthesis. Experimental results demonstrate that our hybrid approach outperforms diffusion-only methods across multiple evaluation metrics and achieves competitive performance compared to state-of-the-art GAN-based methods. The framework successfully generates realistic and geometrically consistent street-view images while preserving fine-grained local details, including street markings, secondary roads, and atmospheric elements such as clouds.

\end{abstract}
\begin{IEEEkeywords}
Street-view synthesis, cross-view translation, satellite images, stable diffusion, hybrid approach.
\end{IEEEkeywords}
\IEEEpeerreviewmaketitle

\section{Introduction}
\IEEEPARstart{T}{he} rapid advancement of computer vision and deep learning has revolutionized our ability to understand and synthesize visual content across diverse domains. Among these developments, satellite-to-street-view synthesis has emerged as a particularly challenging and impactful area of research within the broader cross-view translation domain~\cite{lu2020}. This task involves generating realistic street-view images that maintain geometric and semantic consistency with their corresponding satellite imagery, bridging the substantial gap between aerial and ground-level perspectives. Street-view imagery has become an indispensable resource for modern urban analytics, facilitating comprehensive geospatial data collection, enabling evidence-based decision-making processes, and providing critical insights into urban infrastructure and development patterns~\cite{biljecki2021street}. The ability to synthesize high-quality street-view content from readily available satellite imagery holds particular significance for enhancing visual documentation and analysis in remote, hazardous, or otherwise inaccessible locations where traditional ground-level data collection is either prohibitively expensive or logistically unfeasible~\cite{lu2020},~\cite{shi2022geometry}.

The task of satellite-to-street-view synthesis presents significant challenges due to the limited visual overlap between satellite and street views, resulting in substantially different visual characteristics and constraining the information available for cross-domain inference. For instance, satellite images typically capture only building rooftops, providing minimal information about other architectural features such as building facades~\cite{shi2022geometry},~\cite{wu2022cross}. Additionally, areas with similar street patterns in satellite imagery may appear distinctly different at street level, creating a one-to-many mapping scenario that contributes to limited diversity in generated street-view images. Furthermore, street-view images tend to display fine-grained local details, including vehicles and pedestrians in urban environments, which are absent in satellite images, thereby complicating the generation of these smaller-scale objects~\cite{lu2020}.

Street-view synthesis serves multiple applications across various domains, particularly in cross-view geo-localization tasks that aim to localize street-view images by matching them against geo-referenced satellite image databases. This capability plays a crucial role in applications such as augmented reality, autonomous driving, event detection, and robotics~\cite{toker2021coming},~\cite{wang2021each}. Since retrieval systems can only recognize and locate similar objects, integrating satellite-to-street-view synthesis becomes essential for enhancing the generator's ability to learn useful information that improves location retrieval accuracy~\cite{toker2021coming}. 
Moreover, street-view data is fundamental for developing sidewalk GIS datasets, which constitute key components in smart city development. To effectively utilize sidewalks in various smart city applications, researchers require publicly accessible sidewalk GIS data that provides comprehensive information about sidewalk dimensions, coverage, location, and physical conditions~\cite{biljecki2021street},~\cite{kang2021developing}. However, many cities lack such publicly available GIS data for sidewalks. Street-view images have been successfully employed to extract sidewalk information in sidewalk detection tasks~\cite{cheng2018curb}, while both satellite and street-view images are utilized in sidewalk extraction tasks~\cite{ning2022sidewalk}. Additionally, street-view images have proven valuable in urban analytics and urban planning domains, where they are used to evaluate urban infrastructure and services~\cite{zund2021street} and detect urban changes~\cite{xia2022duarus},~\cite{lyu2023large},~\cite{byun2022street}.

Previous research has extensively investigated satellite-to-street-view synthesis using Generative Adversarial Network (GAN)-based methods~\cite{shi2022geometry},~\cite{toker2021coming},~\cite{deng2018like},~\cite{ren2021cascaded}. Additional approaches have explored the integration of Bicycle GAN and U-Net architectures~\cite{lu2020}. Other studies have focused on conditional GANs that simultaneously learn to generate both satellite and street-view images alongside their corresponding segmentation maps~\cite{wu2022cross_prog},~\cite{regmi2018cross}. Furthermore, semantic-guided street-view generation has been explored using conditional GANs, where semantic maps are concatenated with input satellite images~\cite{tang2020local},~\cite{tang2019multi}. Conversely, street-view to satellite image synthesis has been investigated using conditional GANs for geo-localization tasks aimed at estimating street-view image locations~\cite{regmi2019bridging}. Additional research has addressed street-view geo-localization through feature-matching methods employing Convolutional Neural Networks (CNNs), Siamese networks, and Triplet networks~\cite{workman2015wide},~\cite{shi2020looking},~\cite{vo2016localizing},~\cite{Lin_2015_CVPR},~\cite{hu2018cvm},~\cite{liu2019lending}.

While satellite imagery accessibility has significantly enhanced our ability to capture images of all Earth locations, street-view images lack such comprehensive coverage~\cite{lu2020}. Vast geographical areas remain sparsely represented by street-view imagery, with the majority of available street-view images concentrated near major roads and well-known landmarks~\cite{workman2015wide}.

Despite significant progress in the satellite-to-street-view synthesis domain, current methodologies face important limitations. Existing approaches predominantly rely on generative adversarial networks and their variants, alongside various CNN architectures for feature extraction and Siamese or Triplet networks for geo-localization tasks. Recently, several studies have explored the application of Stable Diffusion models for satellite-to-street-view synthesis, representing a significant advancement in terms of image quality and detailed generation capabilities. However, some generated results exhibit geometric inconsistencies and introduce additional structural elements that are absent in ground-truth images, resulting in misplaced components, such as sidewalks, and inaccurate road layouts. 

To address these limitations, we propose an innovative hybrid framework that integrates the strengths of Stable Diffusion and GAN models to enhance both geometric consistency and realism in generated images. The main contributions of this work are summarized as follows:

\begin{itemize}
    \item We propose a hybrid approach that combines the strengths of both Stable Diffusion models and Generative Adversarial Networks for generating panoramic street-view images with improved geometric consistency.

    \item Our hybrid framework demonstrates significant improvements over existing methods, outperforming the most recent GAN-based approaches on the CVUSA dataset by 2.18\% for SSIM, 2.68\% for FID, and 11.75\% in PSNR compared to pure diffusion-based methods.
\end{itemize}
\section{Related work}
This section reviews the relevant literature across four key areas: image-to-image translation, conditional generative adversarial networks, geometric correspondence methods, and representation-based approaches for cross-view synthesis and matching.

\noindent\textbf{Image-to-Image Translation.} Image-to-image translation aims to learn mappings between input and output image domains through supervised learning using paired training data. Isola~\emph{et~al.}~\cite{isola2017} introduced the seminal pix2pix framework, which uses a conditional GAN to learn translation functions from input to output image domains. Building upon this foundation, pix2pixHD~\cite{wang2018high} enhanced the original pix2pix approach by incorporating a multi-scale discriminator and a coarse-to-fine generator architecture, enabling high-resolution image synthesis. In contrast to paired training approaches, Zhu~\emph{et~al.}~\cite{zhu2017unpaired} proposed CycleGAN, which performs image-to-image translation using unpaired training data through cycle-consistency constraints.

\noindent\textbf{Conditional Generative Adversarial Networks (cGAN).} Conditional GANs extend the foundational work of Goodfellow~\emph{et~al.}~\cite{goodfellow2014generative} by incorporating additional conditional inputs to enable controlled image synthesis. Mirza and Osindero~\cite{mirza2014conditional} first introduced the conditional GAN framework, allowing generators to produce user-specified image outputs. In the context of cross-view synthesis, Regmi~\emph{et~al.}~\cite{regmi2018cross} proposed X-Fork and X-Seq cGAN-based models for bidirectional satellite-to-street view synthesis, incorporating corresponding segmentation maps to improve spatial consistency. Deng~\emph{et~al.}~\cite{deng2018} investigated the utility of cGAN-generated street-view images for land-cover classification tasks, demonstrating the practical applications of synthetic imagery.
Further advancing this line of work, Regmi~\emph{et~al.}~\cite{regmi2019bridging} developed a street-view to satellite image framework that leverages learned representations from cGAN-generated satellite images for cross-view matching tasks. Lu~\emph{et~al.}~\cite{lu2020} introduced a geo-transformation module that generates geometrically consistent street-view images by utilizing segmentation and depth information extracted from satellite imagery, addressing the critical challenge of geometric consistency in cross-view synthesis.
Several frameworks have explored multi-generator and multi-discriminator architectures to bridge the substantial viewpoint gap in cross-view translation. Zhu~\emph{et~al.}~\cite{zhu2018generative} introduced an intermediate homography view based on multi-GAN models for synthesizing frontal-to-bird's-eye view transformations. Tang~\emph{et~al.}~\cite{tang2019multi} investigated semantic-guided street-view generation through a multi-channel module designed to select optimal intermediate generations for enhanced output quality. Similarly, Tang~\emph{et~al.}~\cite{Tang_2020_CVPR} examined semantic-guided generation using a joint generation network that leverages both local class-specific and global image-level features through dual discriminators and fusion attention mechanisms.
Wu~\emph{et~al.}~\cite{wu2022cross} proposed a framework that simultaneously generates street-view images and their corresponding segmentation maps from satellite inputs, implementing cross-stage attention mechanisms across multiple generation stages. More recently, Qian~\emph{et~al.}~\cite{qian2023sat2density} presented a GAN-based framework that learns 3D geometry through density representations, eliminating the requirement for explicit depth information in street-view image generation.

\noindent\textbf{Geometric Correspondence.} Maintaining geometric consistency between satellite and street-view imagery is crucial for preserving spatial relationships during cross-view synthesis and matching tasks. Hu~\emph{et~al.}~\cite{hu2018cvm} developed a Siamese-based framework incorporating NetVLAD layers to jointly learn robust representations specifically tailored for cross-view matching applications. Vo~\emph{et~al.}~\cite{vo2016localizing} introduced a framework featuring a novel distance-based logistic (DBL) layer designed to enhance Siamese and Triplet network performance in geo-localization tasks.
Building upon these approaches, Liu~\emph{et~al.}~\cite{Liu_2019_CVPR} presented an enhanced Siamese-based model that incorporates orientation geometry information to simultaneously learn from image appearance features. Shi~\emph{et~al.}~\cite{shi2020looking} developed a Dynamic Similarity Matching (DSM) network specifically designed to evaluate feature similarity between paired images while considering street-view image orientation. Toker~\emph{et~al.}~\cite{toker2021coming} introduced a multi-task framework that simultaneously handles retrieval and cross-view synthesis tasks, enabling models to learn feature representations that enhance retrieval capabilities while generating images across domains. Most recently, Shi~\emph{et~al.}~\cite{shi2022geometry} developed a satellite-to-street-view projection (S2SP) module that learns to generate height maps and translates satellite images into geometrically consistent street-view imagery.

\noindent\textbf{Representation-Based Approaches.} These methods focus on extracting meaningful features that capture semantic information from both satellite and street-view images for various downstream applications, particularly cross-view matching. Workman~\emph{et~al.}~\cite{workman2015wide} proposed a CNN-based architecture that effectively learns to integrate semantic features from both satellite and street-view domains, addressing inherent challenges in cross-view matching tasks. Zhai~\emph{et~al.}~\cite{zhai2017} addressed street-view geo-localization and geo-orientation by developing a model that predicts semantic segmentations of street-view images using aligned satellite imagery, facilitating semantic segmentation transfer between domains. Lin~\emph{et~al.}~\cite{Lin_2015_CVPR} presented a Siamese-based model that learns low-dimensional feature representations with the objective of minimizing distances between images from identical locations while maximizing distances between images from different locations.
Despite these significant advances, existing approaches primarily rely on GAN-based architectures and traditional CNN features, with limited exploration of diffusion-based methods for cross-view synthesis. Recent developments in diffusion models, particularly Stable Diffusion, have demonstrated remarkable capabilities in high-quality image generation, motivating our investigation into hybrid approaches that combine the strengths of both diffusion and GAN-based methods for improved cross-view synthesis.

\section{Methodology}\label{methods}
\begin{figure*}[t]
    \centering
    \includegraphics[width=\textwidth]{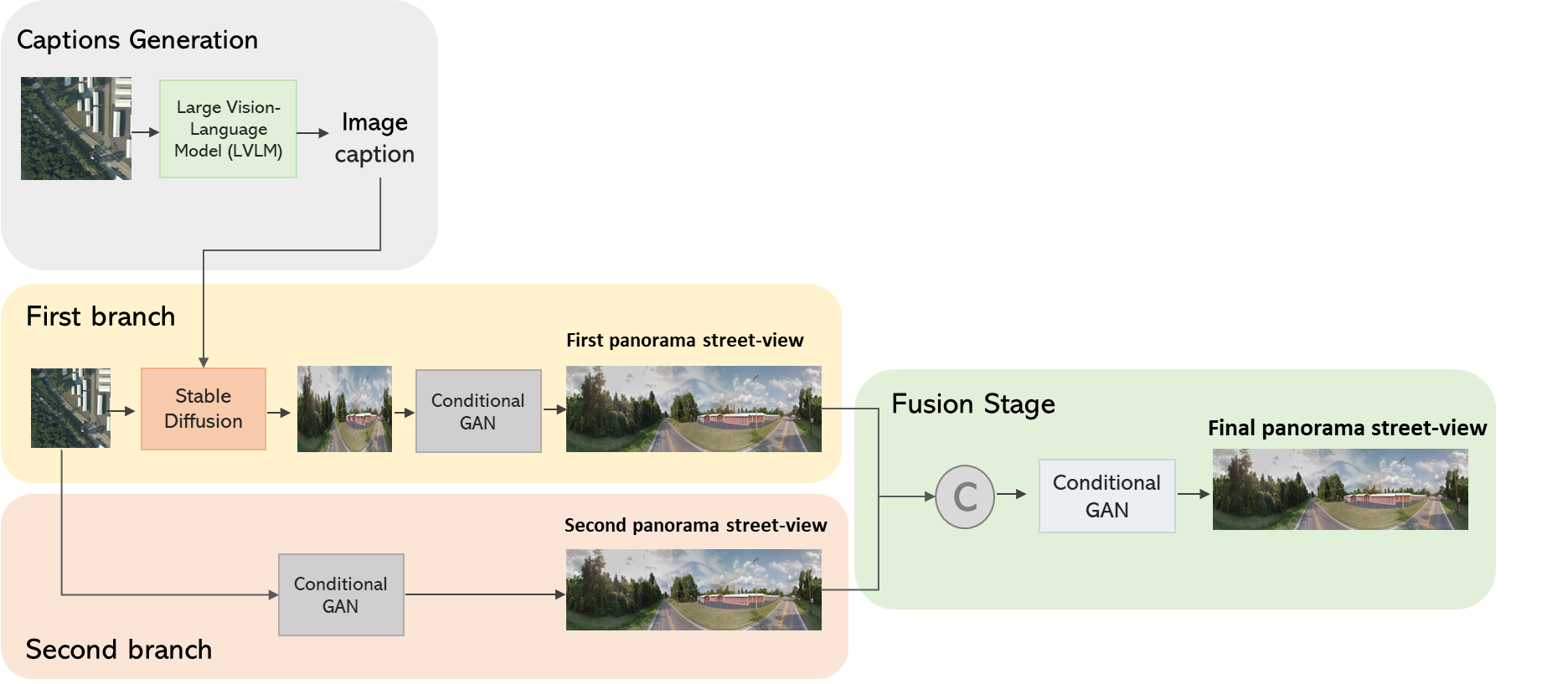}
    \caption{Overview of our proposed hybrid framework for satellite-to-street-view synthesis.}
    \label{fig:sat2street}
\end{figure*}

This section presents our hybrid framework for satellite-to-street-view synthesis that combines the strengths of diffusion-based models and generative adversarial networks. Our approach leverages a dual-branch architecture to learn complementary representations of synthesized street-view images. By fusing the representations from both branches, we construct a robust representation that enhances the generation of geometrically consistent and realistic street-view outputs. Figure \ref{fig:sat2street} provides an overview of our proposed framework.

\subsection{Stable Diffusion Model}

The Stable Diffusion (SD) model~\cite{rombach2022high} is a probabilistic text-to-image generative model designed to generate images $x_{0}$ by progressively denoising random Gaussian noise $x_{t}$ over $T$ timesteps. The SD model operates through two distinct phases: the forward diffusion process and the reverse denoising process.

The forward diffusion process consists of multiple steps in latent space where Gaussian noise is iteratively added to each training image. The noise amount varies at each step until the image becomes pure Gaussian noise~\cite{croitoru2023diffusion}. Given an original training sample $x_0 \sim p(x_0)$, noisy versions $x_1, x_2, x_3, \ldots, x_t$ are generated according to the Markov chain process described in Eq.~\ref{eq:forward}:

\begin{equation}
p(x_{t}|x_{t-1})=\mathcal{N}(x_{t};\sqrt{1-\beta_{t}} \cdot x_{t-1},\beta_{t} \cdot \mathbf{I}), \forall t \in \{1,\ldots,T\},
\label{eq:forward}
\end{equation}

where $\beta_{1}, \beta_{2}, \ldots, \beta_{t} \in [0, 1)$ are hyperparameters that define the variance schedule over multiple diffusion steps, $T$ represents the total number of diffusion steps, $\mathcal{N}(x;\mu,\sigma)$ denotes the Gaussian distribution with mean $\mu$ and variance $\sigma$, and $\mathbf{I}$ represents the identity matrix with the same dimensions as the input image $x_{0}$~\cite{croitoru2023diffusion}.

Conversely, the reverse denoising process enables image generation from $p(x_{0})$ by starting with completely random noise $x_{t} \sim \mathcal{N}(0,\mathbf{I})$ and applying the reverse denoising process~\cite{croitoru2023diffusion} as described in Eq.~\ref{eq:reverse}:

\begin{equation}
p(x_{t-1}|x_{t})=\mathcal{N}(x_{t-1};\mu(x_{t},t),\Sigma(x_{t},t))
\label{eq:reverse}
\end{equation}

These steps can be approximated by training a neural network, such as U-Net~\cite{ronneberger2015u}, to predict the added noise $\epsilon$ given the noisy image $x_{t}$ and timestep $t$~\cite{croitoru2023diffusion}. By applying the reverse denoising process for all timesteps $T$, we can recover the data distribution from a purely noisy image $x_{t}$.

The SD model~\cite{rombach2022high} comprises multiple interconnected components: a Contrastive Language-Image Pre-Training (CLIP)~\cite{radford2021learning} text encoder, a U-Net~\cite{ronneberger2015u} architecture, and a Variational Autoencoder (VAE)~\cite{kingma2014auto}.

The U-Net~\cite{ronneberger2015u} is a convolutional neural network architecture consisting of a contracting path and an expansive path, which perform downsampling and upsampling operations, respectively. The contracting path follows a conventional convolutional network structure, comprising repeated applications of two unpadded $3 \times 3$ convolution layers. Each convolution is followed by a Rectified Linear Unit (ReLU) activation function and downsampling through a $2 \times 2$ max pooling operation with stride 2. Additionally, the number of feature channels doubles at each downsampling step~\cite{ronneberger2015u}. 

The expansive path involves upsampling the feature maps through a $2 \times 2$ up-convolution that halves the number of channels~\cite{ronneberger2015u}. The upsampled feature map is then concatenated with the corresponding cropped features from the contracting path, followed by two $3 \times 3$ convolutional layers, each with ReLU activation. In the final layer, a $1 \times 1$ convolution transforms the 64-component feature vector into the desired number of output classes~\cite{ronneberger2015u}.

In the SD model, the U-Net~\cite{ronneberger2015u} serves as the denoising network. It takes the noisy image $x_{t}$ and current timestep $t$ to predict the noise, which is then subtracted to obtain a less noisy image~\cite{croitoru2023diffusion},~\cite{rombach2022high}. This process is performed progressively over multiple timesteps.

The VAE~\cite{kingma2014auto} is a likelihood-based generative model consisting of an encoder and decoder. The encoder transforms an input image $x$ into a latent representation $z$, while the decoder reconstructs the image $\hat{x}$ from the latent representation $z$~\cite{kingma2014auto}.

\subsection{ControlNet}

\begin{figure}[tbp]
    \centering
    \includegraphics[width=9cm]{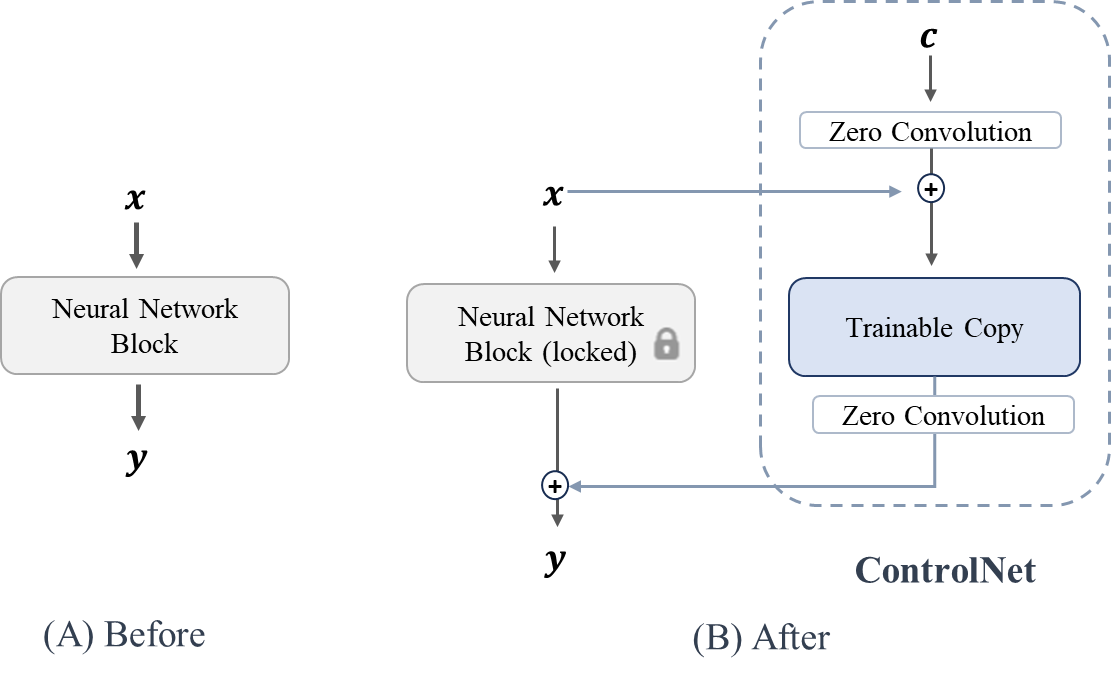}
    \caption{ControlNet integration: (a) original neural network block, (b) ControlNet-enhanced block with additional conditioning.}
    \label{fig:controlnet}
\end{figure}

ControlNet~\cite{zhang2023adding} is a neural network architecture designed to enhance pretrained diffusion models by incorporating additional conditioning controls. When integrating ControlNet into a pretrained neural network block, as illustrated in Figure~\ref{fig:controlnet}, the original block's parameters $\Theta$ are frozen while creating a trainable copy with parameters $\Theta_{c}$~\cite{zhang2023adding}. These components are connected using zero convolutional layers, denoted as $\mathcal{Z}(\cdot;\cdot)$, which minimize harmful noise during training. This connection is implemented as $1 \times 1$ convolution layers with weights and biases initialized to zero.

An additional conditioning vector $c$ is fed into the trainable copy, enabling control over the pretrained diffusion model~\cite{zhang2023adding}, as shown in Figure~\ref{fig:controlnet}b.

Consider a trained neural network $\mathcal{F}(\cdot; \Theta)$ with parameters $\Theta$ that transforms input feature map $x$ into output feature map $y$~\cite{zhang2023adding}:

\begin{equation}
y = \mathcal{F}(x;\Theta)
\label{eq:nn}
\end{equation}

To construct a ControlNet, we utilize two zero convolution instances with parameters $\Theta_{z1}$ and $\Theta_{z2}$~\cite{zhang2023adding}. The ControlNet output $y_{c}$ is calculated as:

\begin{equation}
y_{c} = \mathcal{F}(x;\Theta) + \mathcal{Z}(\mathcal{F}(x+\mathcal{Z}(c;\Theta_{z1});\Theta_{c});\Theta_{z2})
\label{eq:controlnett}
\end{equation}

During training initialization, the zero initialization of weights and biases in the zero convolution layers causes both $\mathcal{Z}(\cdot;\cdot)$ terms in Eq.~\ref{eq:controlnett} to evaluate to zero, resulting in $y_{c} = y$.

\begin{figure}[tbp]
    \centering
    \includegraphics[width=9cm]{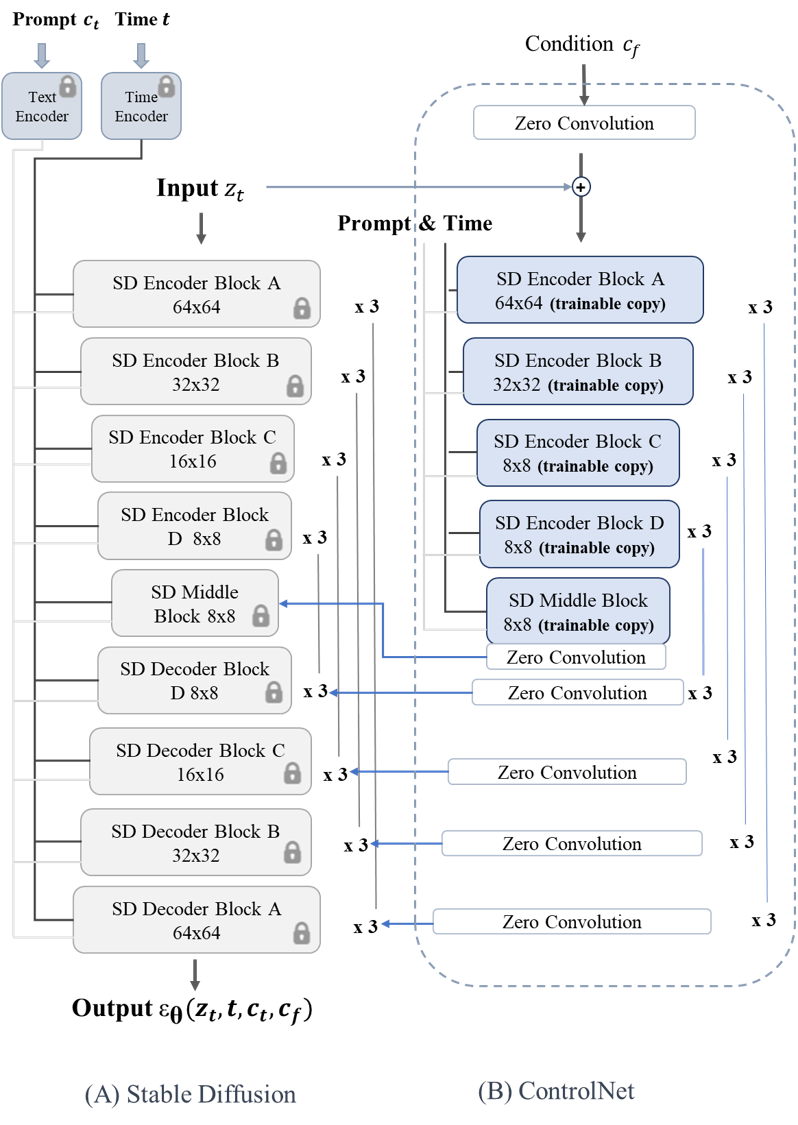}
    \caption{Integration of ControlNet with Stable Diffusion architecture.}
    \label{fig:cn+sd}
\end{figure}

ControlNet~\cite{zhang2023adding} is integrated with the U-Net~\cite{ronneberger2015u} architecture of Stable Diffusion at the encoder blocks and middle block, as illustrated in Figure~\ref{fig:cn+sd}. Specifically, it creates trainable copies of Stable Diffusion's 12 encoding blocks and 1 middle block. These encoding blocks operate at four resolutions: $64 \times 64$, $32 \times 32$, $16 \times 16$, and $8 \times 8$, with each encoder block replicated three times. The resulting outputs are added to the U-Net's 12 skip connections and 1 middle block. Timestep encoding $t$ is performed using a time encoder with positional encoding~\cite{zhang2023adding}.

\subsection{Conditional Generative Adversarial Network (cGAN)}

Generative Adversarial Networks (GANs)~\cite{goodfellow2014generative} represent an innovative approach for training generative models through adversarial learning. A GAN consists of two competing neural networks: a Generator $G$ and a Discriminator $D$.

The Generator's primary objective is to deceive the Discriminator by generating synthetic data that resembles the true data distribution. The Generator constructs a mapping function from a prior noise distribution $p_{z}(z)$ to the data space, learning the generator distribution $p_{g}$ over data $x$, denoted as $G(z;\theta_{g})$~\cite{goodfellow2014generative},~\cite{mirza2014conditional}. The Discriminator, denoted as $D(x; \theta_{d})$, outputs a scalar probability indicating whether sample $x$ originated from the training data rather than from $p_{g}$~\cite{goodfellow2014generative},~\cite{mirza2014conditional}. 

Both models are trained jointly in a two-player minimax game with value function $V(D,G)$:

\begin{multline}
\min_{G} \max_{D} V(D,G) = \mathbb{E}_{x \sim p_{\text{data}}(x)}[\log D(x)] + \\ \mathbb{E}_{z \sim p_{z}(z)}[\log(1-D(G(z)))]
\label{eq:gan}  
\end{multline}

The Generator training minimizes the $\log(1 - D(G(z)))$ term in Eq.~\ref{eq:gan}, while the Discriminator training maximizes the likelihood of correctly classifying both real training data and synthetic samples generated by $G$~\cite{goodfellow2014generative}.

GANs can be extended to conditional models by incorporating auxiliary information $y$, such as class labels or other data modalities. Both Generator and Discriminator are conditioned on this additional information by incorporating $y$ as an additional input~\cite{mirza2014conditional}. The conditional GAN objective function becomes:

\begin{multline}
\min_{G} \max_{D} V(D,G) = \mathbb{E}_{x \sim p_{\text{data}}(x)}[\log D(x|y)] + \\ \mathbb{E}_{z \sim p_{z}(z)}[\log(1-D(G(z|y)))]
\label{eq:cgan}
\end{multline}

This conditional framework enables controlled generation based on specific input conditions, making it particularly suitable for cross-view synthesis tasks where satellite images serve as conditioning inputs for street-view generation.

\subsection{Implementation Details}
Our proposed framework employs a multi-stage training approach consisting of two parallel branches and a fusion stage, as illustrated in Figure~\ref{fig:sat2street}. This architecture leverages the complementary strengths of diffusion-based and GAN-based models to achieve superior geometric consistency and visual quality in cross-view synthesis.
\subsubsection{Branch 1: Diffusion-Based Pipeline}
The first branch utilizes a diffusion-based approach built upon the Stable Diffusion (SD) model v2.1. We utilize ControlNet~\cite{zhang2023adding} as the backbone architecture for fine-tuning the SD model, allowing for precise control over the generation process through satellite image conditioning. The ControlNet integration allows the diffusion model to maintain spatial relationships and geometric consistency between the input satellite imagery and the generated street-view content.
Following the initial generation by the SD model, the output requires geometric transformation to match the panoramic street-view format. To address this, we integrate PanoGAN~\cite{wu2022cross} as a conditional GAN-based backbone that transforms the square street-view images generated by the SD model into the panoramic format. This component functions as a specialized resizing and format conversion module, ensuring that the diffusion-generated content conforms to the expected panoramic street-view dimensions while preserving the spatial relationships established during the diffusion process.
\subsubsection{Branch 2: Direct GAN-Based Pipeline}
The second branch implements a direct satellite-to-street-view translation approach using PanoGAN~\cite{wu2022cross}, which is applied directly to the input satellite images. This branch serves as a complementary pathway that captures different aspects of the cross-view translation task, with a particular focus on the direct mapping relationships between satellite and street-view domains, which GANs excel at learning.
The direct application of PanoGAN enables the model to learn domain-specific features and translation patterns that may be challenging for diffusion models to capture, particularly in terms of local details and texture patterns characteristic of street-view imagery. This branch contributes essential geometric and contextual information that enhances the overall synthesis quality.
\subsubsection{Fusion Stage}
The fusion stage combines the outputs from both branches. We concatenate the outputs from the diffusion-based branch (Branch 1) and the GAN-based branch (Branch 2), creating a rich feature representation that incorporates the strengths of both approaches. This concatenated representation is then fed as input to a conditional GAN-based fusion network, which we refer to as the "fusion network" throughout this paper.
The fusion network, also based on the PanoGAN~\cite{wu2022cross} architecture, is specifically trained to synthesize the complementary information from both branches into coherent, high-quality street-view images. This network learns to selectively combine features from each branch, leveraging the detailed texture generation capabilities of the diffusion model while maintaining the geometric consistency and spatial relationships learned by the direct GAN approach.
\subsubsection{Training Strategy}
The multi-stage training approach ensures optimal performance at each component level:
\begin{itemize}
    \item Stage 1: Fine-tune the ControlNet-enhanced Stable Diffusion model using satellite-street-view pairs, focusing on learning the fundamental cross-view relationships and spatial correspondences.
    \item Stage 2: Train the panoramic conversion module to transform the SD outputs into a proper panoramic format while maintaining the quality of generated content.
    \item Stage 3: Train the direct PanoGAN branch on the same dataset to learn complementary translation patterns and domain-specific features.
    \item Stage 4: Train the fusion network to optimally combine outputs from both branches, learning to leverage the strengths of each approach while mitigating their limitations.
\end{itemize}
This sequential training strategy allows each component to specialize in its designated function while ensuring compatibility and effective integration during the fusion stage. The resulting framework demonstrates improved geometric consistency, enhanced detail preservation, and superior overall visual quality compared to single-model approaches.

\section{Experiments}\label{experiments}

\subsection{Experimental Settings}

\begin{figure*}[tbp]
  \hspace{-1cm}
  \begin{tabular}{cc}
    \includegraphics[height=2cm, width=8.5cm]{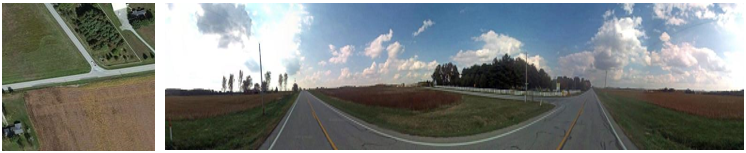}& 
    \includegraphics[height=2cm, width=8.5cm]{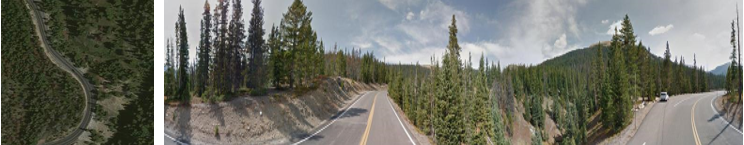}\\
    \includegraphics[height=2cm, width=8.5cm]{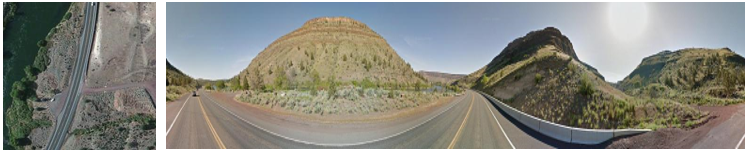}&
    \includegraphics[height=2cm, width=8.5cm]{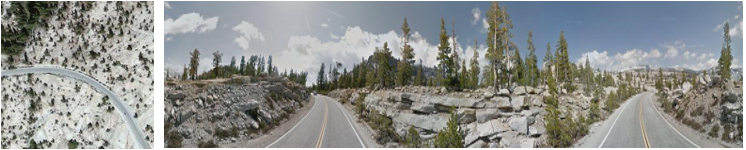} \\
  \end{tabular}
    \caption{Representative satellite and street-view image pairs from the CVUSA dataset, demonstrating the diversity of urban and rural environments.}
   \label{fig:CVUSA}
\end{figure*}

\noindent\textbf{Dataset.} We evaluate our proposed framework on the CVUSA benchmark dataset, a widely-used standard for cross-view synthesis tasks. The dataset contains 35,532 pairs of satellite and panoramic street-view images for training and 8,884 pairs for testing. CVUSA images capture diverse street scenes from both urban and rural areas across the United States, with satellite images at $750 \times 750$ resolution and street-view images at $1232 \times 224$ resolution. Figure~\ref{fig:CVUSA} presents representative sample pairs from the dataset, illustrating the diverse geographical and architectural variations present in the data.

\noindent\textbf{Parameter Settings.} Our implementation consists of three main components, each with optimized hyperparameters to ensure effective training and convergence.

\textit{Diffusion Model Configuration:} For the Stable Diffusion component, both satellite and street-view images are resized to $512 \times 512$ resolution to maintain consistency with the pretrained model requirements. Training is conducted for 30 epochs with a batch size of 4, constrained by GPU memory limitations. We employ a cosine learning rate scheduler with an initial learning rate of $1 \times 10^{-3}$. Optimization is performed using AdamW~\cite{loshchilov2017decoupled} with momentum parameters $\beta_{1} = 0.9$ and $\beta_{2} = 0.999$. The DDIM sampler utilizes 50 timesteps during training, with the SD decoder parameters frozen to preserve pretrained capabilities while allowing adaptation through ControlNet.

\textit{PanoGAN Model Configuration:} For the PanoGAN component~\cite{wu2022cross}, satellite images are resized to $256 \times 256$ while street-view images are resized to $256 \times 1024$ to accommodate the panoramic aspect ratio. Training is conducted for 30 epochs with a batch size of 8. The learning rate is set to $1 \times 10^{-3}$ using a cosine learning rate scheduler and Adam optimizer. We adopt the remaining hyperparameters from the original PanoGAN implementation~\cite{wu2022cross} to ensure fair comparison and optimal performance.

\textit{Fusion Network Configuration:} The fusion network combines outputs from both branches through concatenation before processing. Training is performed on the CVUSA dataset for 55 epochs with a batch size of 4. We employ different learning rates for the generator ($1 \times 10^{-4}$) and discriminator ($1 \times 10^{-5}$) to ensure stable adversarial training dynamics. The cosine learning rate scheduler and Adam optimizer are used consistently across all components.

\noindent\textbf{Implementation Details.} The complete framework is implemented in PyTorch. All experiments are conducted on an NVIDIA GeForce RTX 3090 GPU with 24GB memory, which provides sufficient computational resources for training the complex multi-stage architecture while enabling reasonable training and inference times. The multi-stage training approach requires careful coordination between components, with each stage building upon the previous one to achieve optimal performance.

\noindent\textbf{Evaluation Metrics.}
To evaluate the generated street-view images, we utilize measures established in the literature \cite{toker2021coming}, \cite{wu2022cross_prog}. Specifically, we use the Fréchet Inception Distance (FID) and Learned Perceptual Image Patch Similarity (LPIPS) for assessment. Additionally, we incorporate pixel-level evaluations, which include the Peak Signal-to-Noise Ratio (PSNR) and the Structural Similarity Index Measure (SSIM).

\subsection{Comparison with Existing Methods}
In this section, we compare our method with existing methods from both quantitative and qualitative perspectives.

\newcommand{\myfont}{\fontsize{7}{14}\selectfont}

\begin{figure*}[htbp]
    
\hspace{-1 cm}
 \resizebox{20cm}{!}{
\renewcommand{\arraystretch}{1.5}

    \begin{tabular}{
    >{\centering\arraybackslash}p{1.8cm}@{\hskip 1pt}
    >{\centering\arraybackslash}p{1.8cm}@{\hskip 1pt}
    >{\centering\arraybackslash}p{1.8cm}@{\hskip 1pt}
    >{\centering\arraybackslash}p{1.8cm}@{\hskip 1pt}
    >{\centering\arraybackslash}p{1.89cm}@{\hskip 1pt}
    >{\centering\arraybackslash}p{11.3cm}@{\hskip 1pt}
    }

    
    & \textbf{\myfont{Satellite}} &   \textbf{\myfont{LGGAN \newline(global)\cite{tang2020local}}} & 
    \textbf{\myfont{Selection~\newline GAN\cite{tang2019multi}}} & 
    \textbf{\makebox[2cm][l]\small \myfont{Ours}} & \textbf{\myfont{Ground Truth}}  \\
   \multirow{-3}{*}{(a)} &  \includegraphics[width=1.75cm,height=1.75cm]{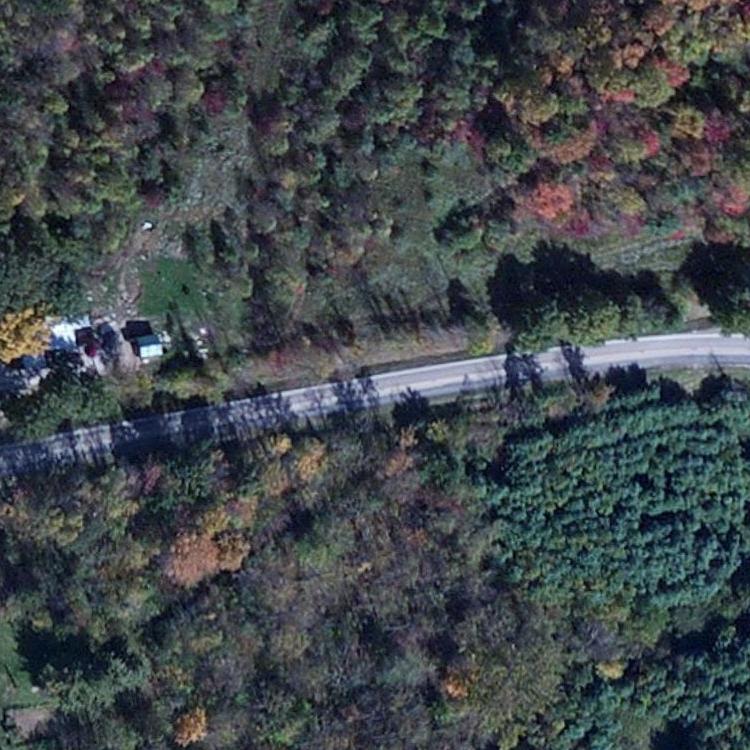}& \includegraphics[width=1.75cm,height=1.75cm]{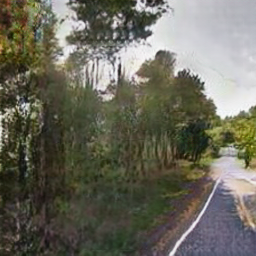} &\includegraphics[width=1.75cm,height=1.75cm]{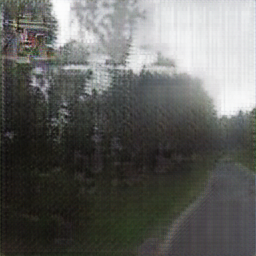} & \includegraphics[width=5cm,height=1.75cm]{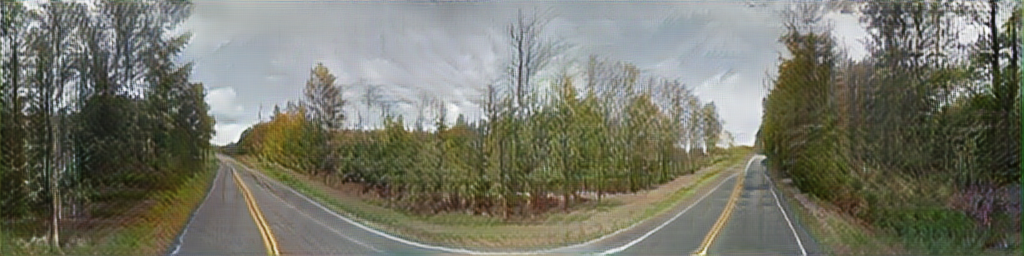}&
   \includegraphics[width=5cm,height=1.75cm]{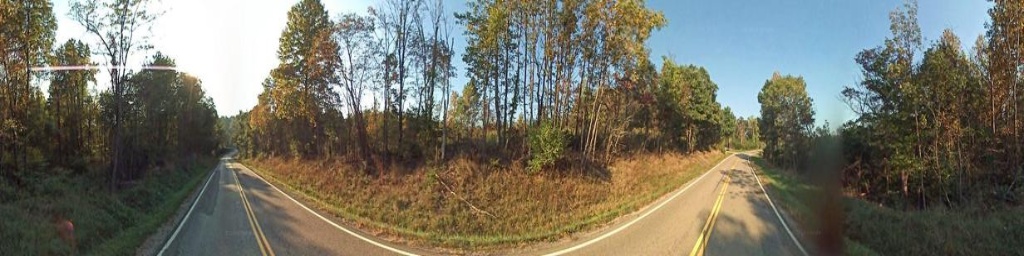} \\
    \multirow{-3}{*}{(b)} & \includegraphics[width=1.75cm,height=1.75cm]{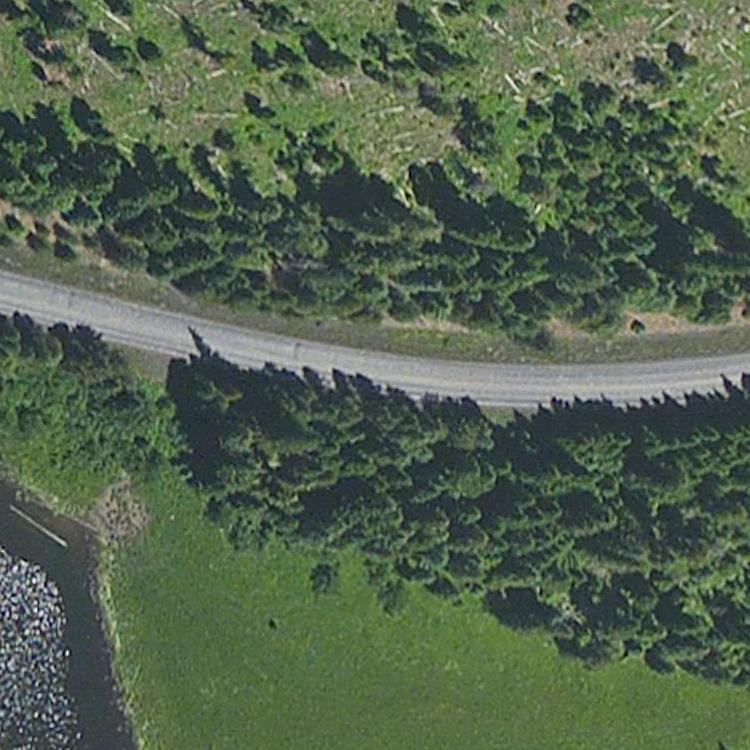} & \includegraphics[width=1.75cm,height=1.75cm]{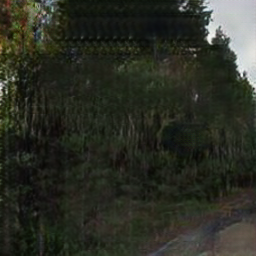} & 
     \includegraphics[width=1.75cm,height=1.75cm]{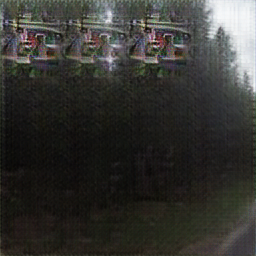} 
     & \includegraphics[width=5cm,height=1.75cm]{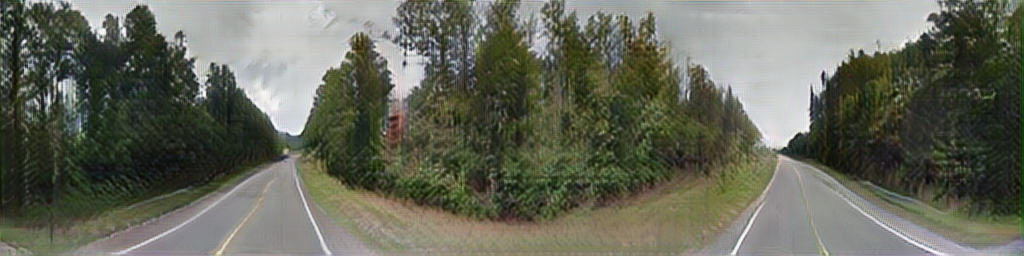} & \includegraphics[width=5cm,height=1.75cm]{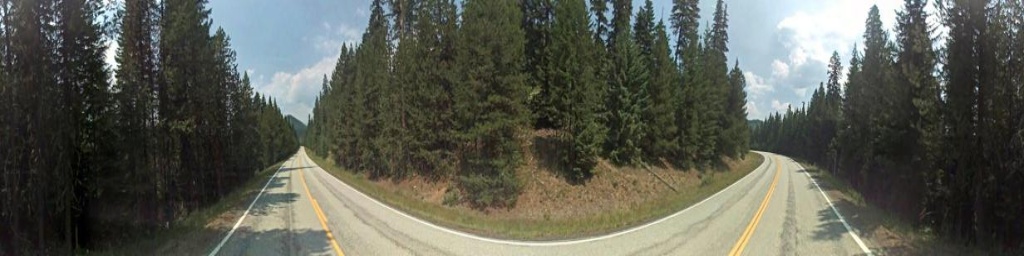}  \\
   \multirow{-3}{*}{(c)} &  \includegraphics[width=1.75cm,height=1.75cm]{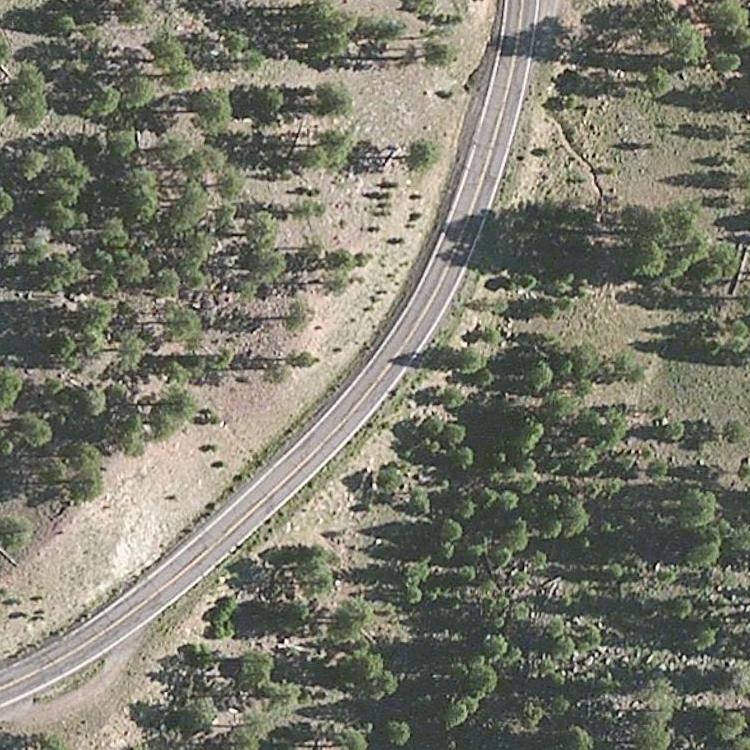} &
   \includegraphics[width=1.75cm,height=1.75cm]{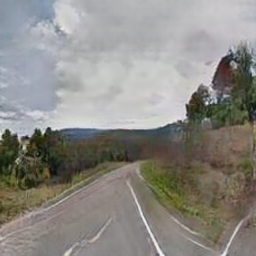}&  \includegraphics[width=1.75cm,height=1.75cm]{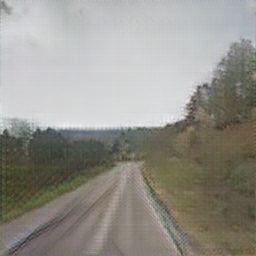} &
   \includegraphics[width=5cm,height=1.75cm]{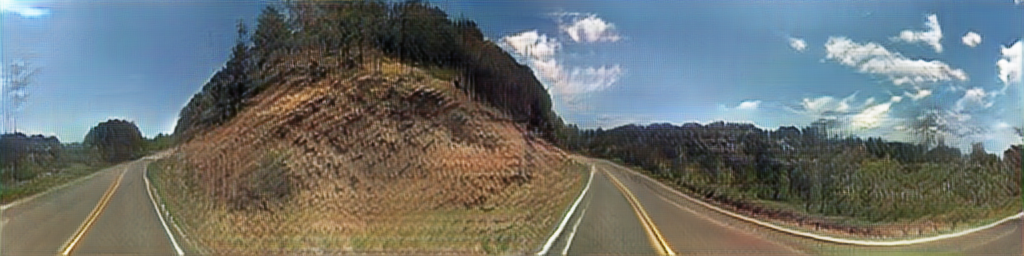} & \includegraphics[width=5cm,height=1.75cm]{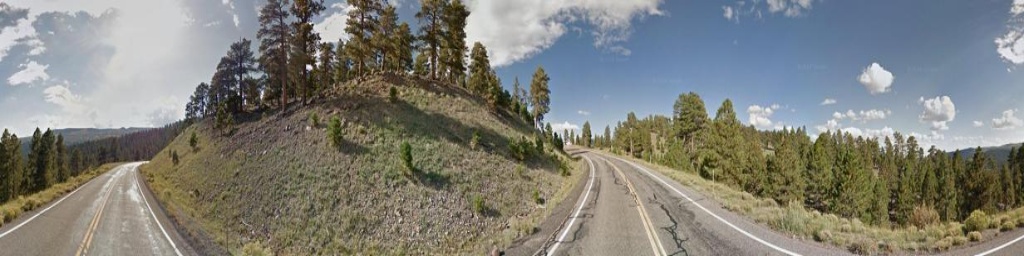} \\
      \multirow{-3}{*}{(d)} & \includegraphics[width=1.75cm,height=1.75cm]{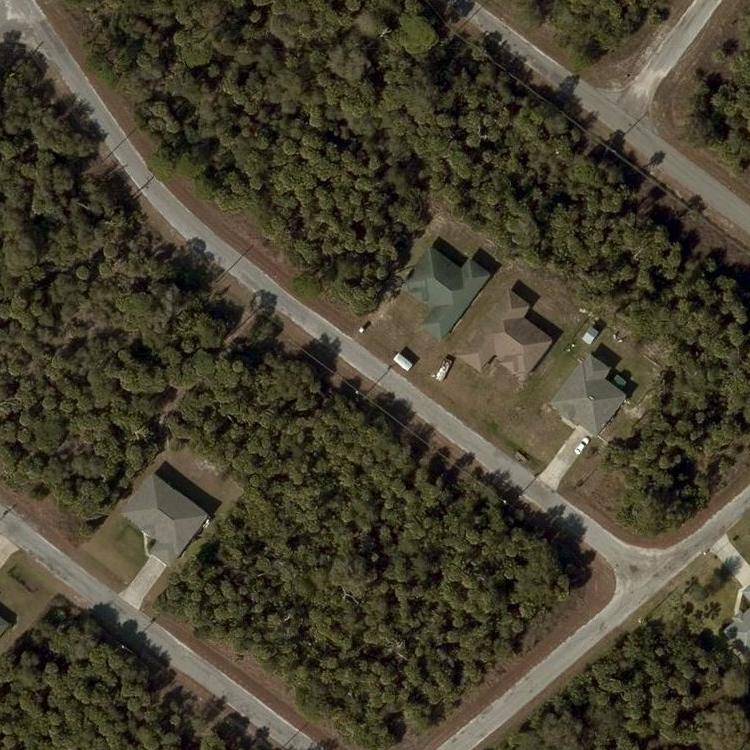} & \includegraphics[width=1.75cm,height=1.75cm]{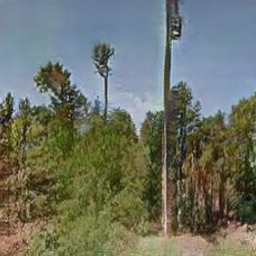}&
       \includegraphics[width=1.75cm,height=1.75cm]{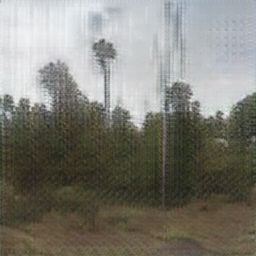} &
      
      \includegraphics[width=5cm,height=1.75cm]{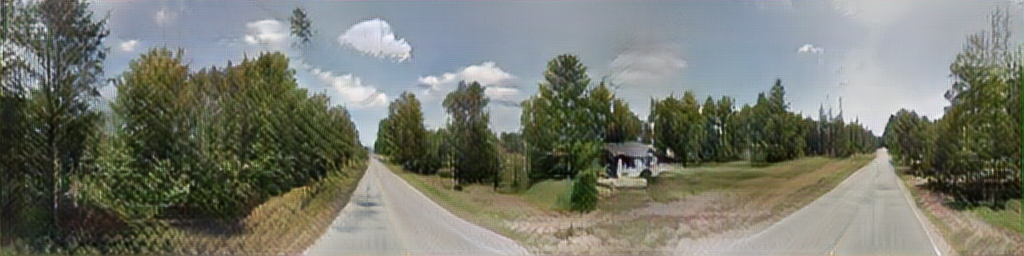}  &\includegraphics[width=5cm,height=1.75cm]{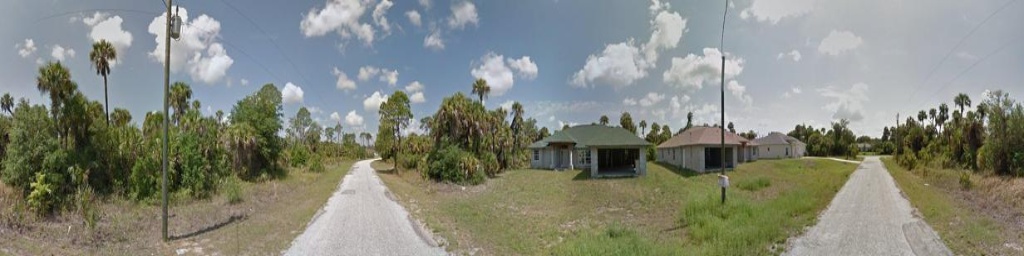} \\
     \multirow{-3}{*}{(e)} &  \includegraphics[width=1.75cm,height=1.75cm]{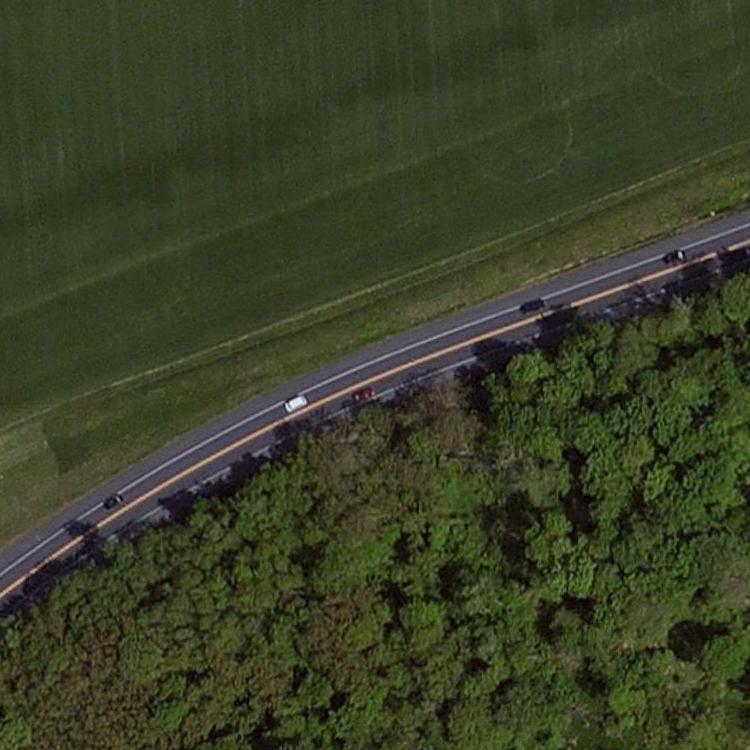} &\includegraphics[width=1.75cm,height=1.75cm]{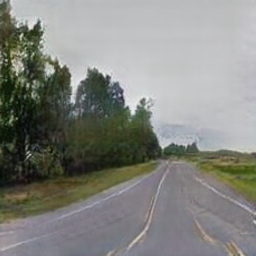} & \includegraphics[width=1.75cm,height=1.75cm]{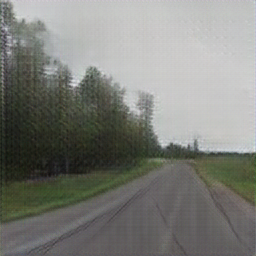}  & \includegraphics[width=5cm,height=1.75cm]{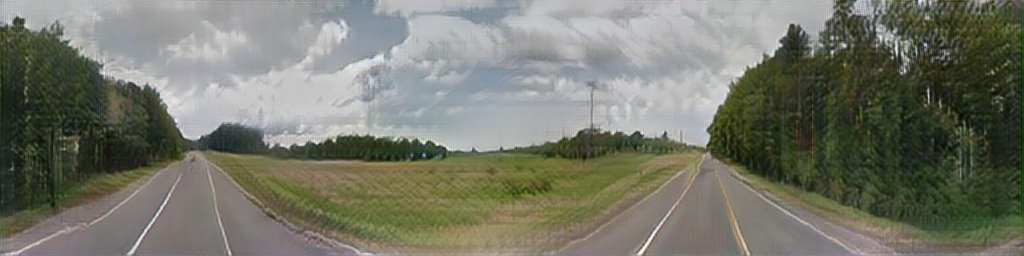} &\includegraphics[width=5cm,height=1.75cm]{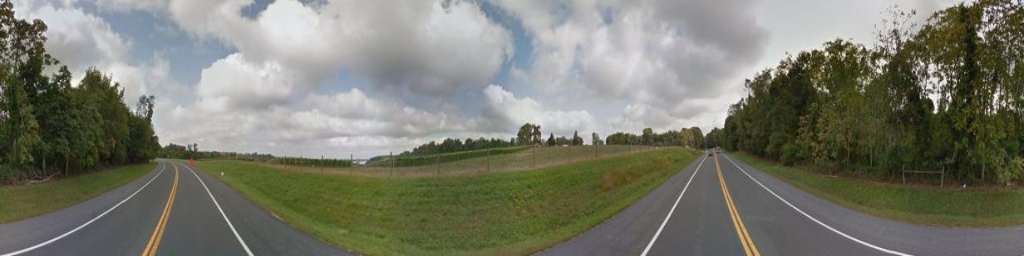} \\

      \multirow{-3}{*}{(f)} &
      \includegraphics[width=1.75cm,height=1.75cm]{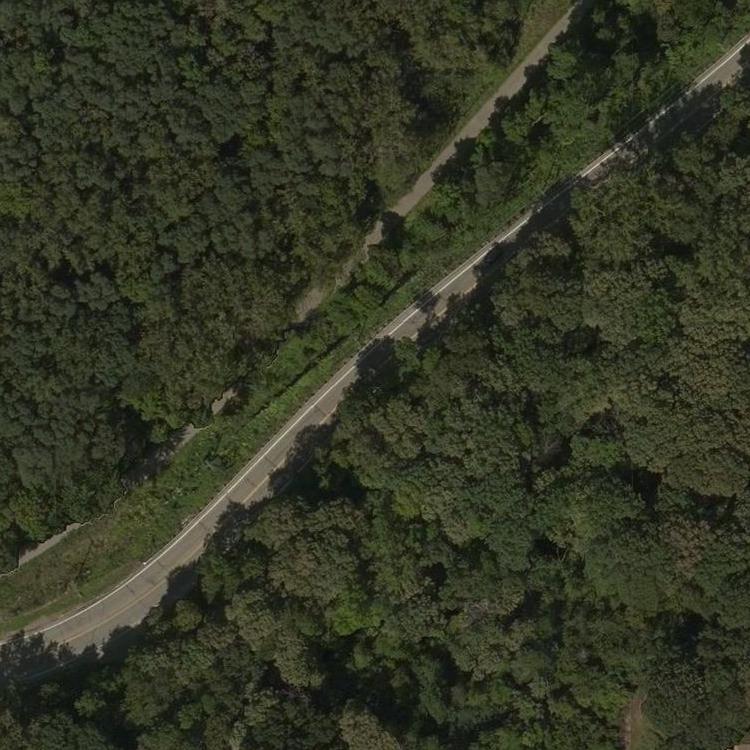} &
      \includegraphics[width=1.75cm,height=1.75cm]{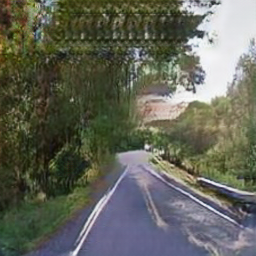}& \includegraphics[width=1.75cm,height=1.75cm]{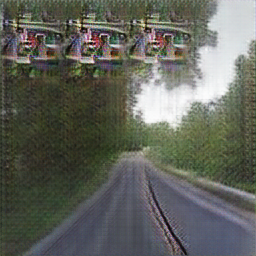}  &
      \includegraphics[width=5cm,height=1.75cm]{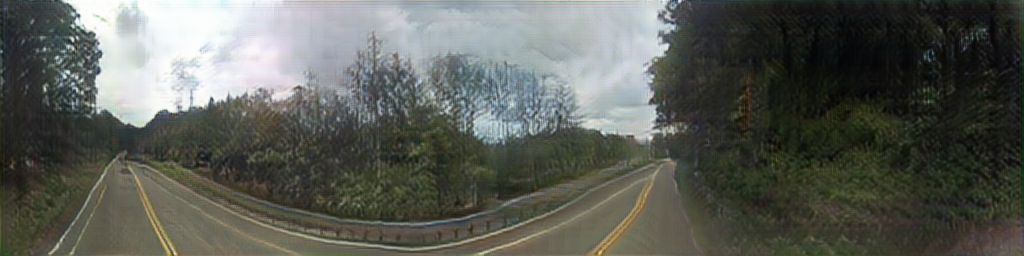}   &
      \includegraphics[width=5cm,height=1.75cm]{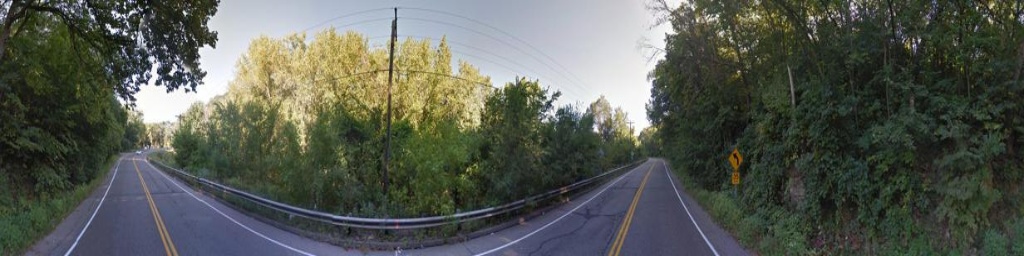} \\
      
       \multirow{-3}{*}{(g)} &
       \includegraphics[width=1.75cm,height=1.75cm]{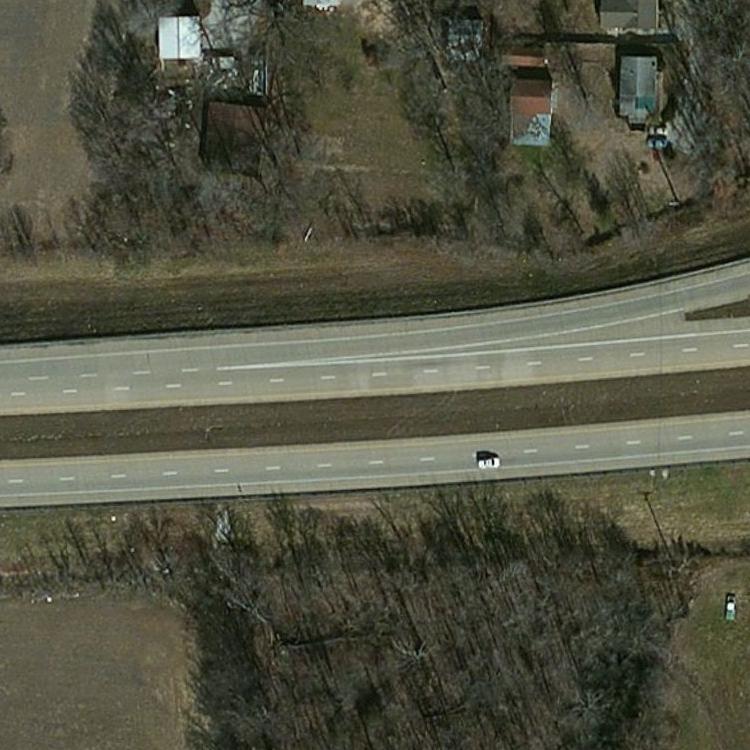} &
       \includegraphics[width=1.75cm,height=1.75cm]{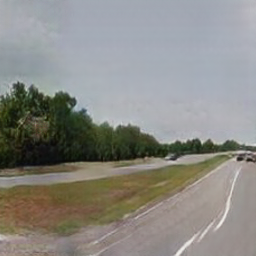}& \includegraphics[width=1.75cm,height=1.75cm]{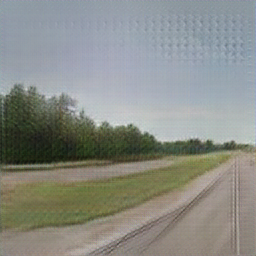}  &
       \includegraphics[width=5cm,height=1.75cm]{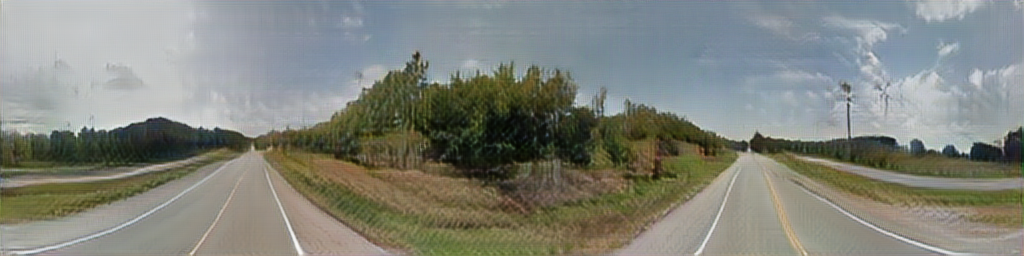} & \includegraphics[width=5cm,height=1.75cm]{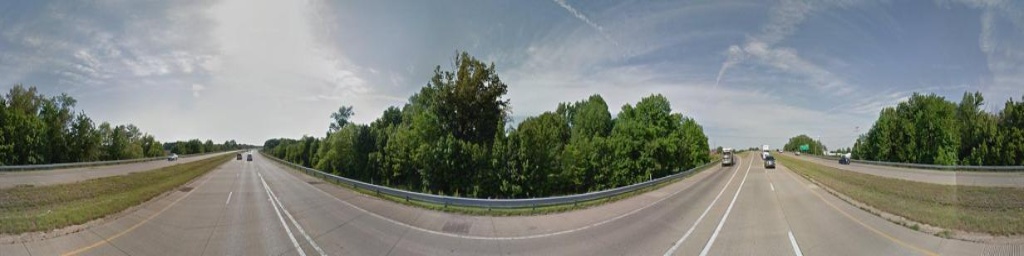} \\
      \multirow{-3}{*}{(h)} &
      \includegraphics[width=1.75cm,height=1.75cm]{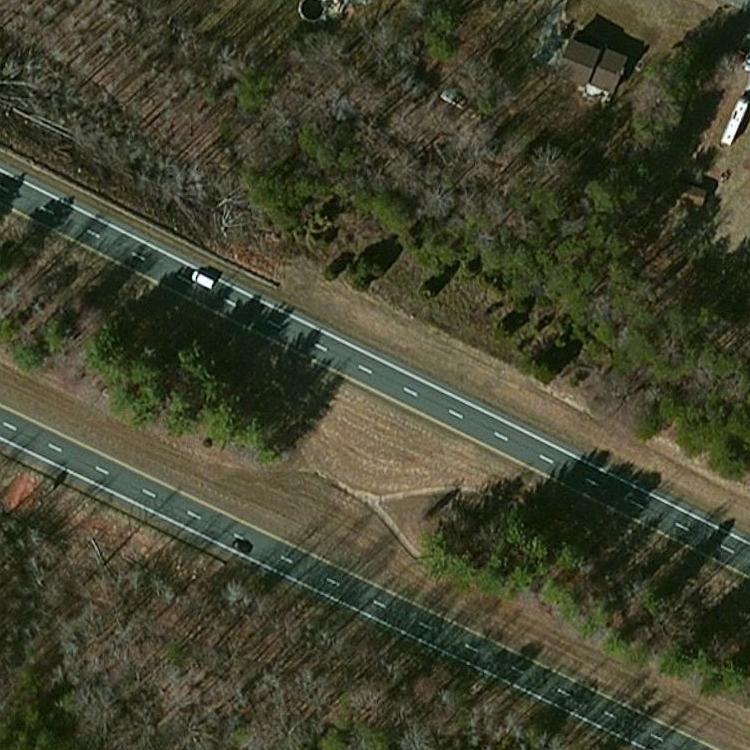} &\includegraphics[width=1.75cm,height=1.75cm]{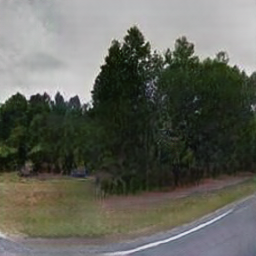} &  \includegraphics[width=1.75cm,height=1.75cm]{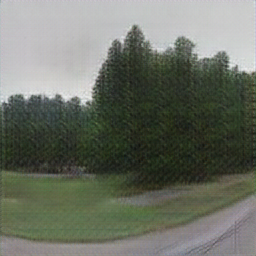}  &
      \includegraphics[width=5cm,height=1.75cm]{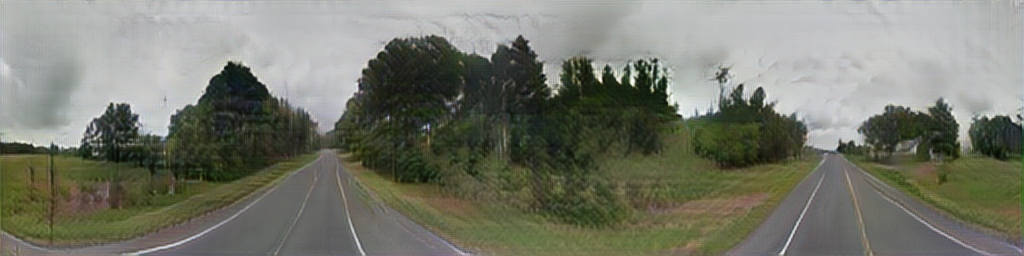}
       &\includegraphics[width=5cm,height=1.75cm]{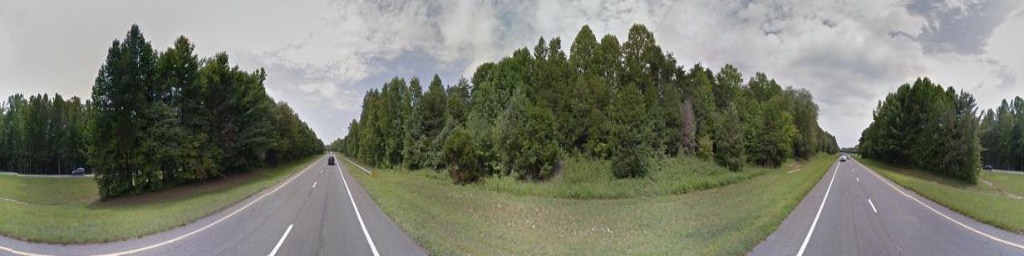} \\

\end{tabular}}
\caption{Qualitative comparison with existing methods.}
\label{fig:qualti-comparision-square}
   
\end{figure*}

\noindent\textbf{Quantitative Comparison.} Table~\ref{tab:Quantitative_comp} presents a comprehensive quantitative comparison of our proposed framework against existing state-of-the-art methods on the CVUSA dataset. Our hybrid approach demonstrates significant improvements across multiple evaluation metrics when compared to both GAN-based and diffusion-only approaches.
Compared to the recent GAN-based method Sat2Density~\cite{qian2023sat2density}, our framework achieves a $2.18\%$ improvement in SSIM score and a $2.68\%$ reduction in FID score. These improvements demonstrate that our hybrid approach generates more realistic images with enhanced structural fidelity compared to purely GAN-based methods. The improved SSIM indicates better preservation of structural information, while the reduced FID score suggests that the generated images more closely match the distribution of real street-view images.
When compared to the diffusion-only approach CrossViewDiff~\cite{li2024crossviewdiff}, our framework achieves a PSNR of 13.41, corresponding to an $11.75\%$ relative improvement. This substantial enhancement highlights the capability of our hybrid approach to generate images with significantly lower pixel-level errors compared to pure diffusion-based methods, demonstrating the effectiveness of combining diffusion and GAN architectures.
However, our framework reports higher LPIPS scores compared to some existing methods. This apparent limitation requires careful interpretation in the context of output format differences. Specifically, methods including Pix2pix, X-Fork, X-Seq, SelectionGAN, PGSL, Cross-MLP, and LGGAN are designed to generate square street-view images ($256 \times 256$), which correspond only to the first quarter of a full panoramic view. In contrast, our framework generates complete $360 ^{\circ}$ panoramic street-view images, presenting a significantly more challenging task for preserving perceptual similarity across the entire panoramic field of view. This fundamental difference in output scope potentially contributes to the higher LPIPS scores and affects SSIM and PSNR comparisons.
While some panorama-generation methods report competitive SSIM and PSNR scores, our hybrid framework introduces a novel paradigm that strategically leverages the complementary strengths of both GANs and diffusion models. This innovative approach presents unique challenges that may impact performance on certain metrics, particularly those designed for evaluation of limited-view synthesis. Nevertheless, the framework demonstrates significant potential for advancing cross-view synthesis, as evidenced by the improved SSIM and FID scores compared to the most recent GAN-based method and the substantial PSNR improvement over pure diffusion-based approaches.
The excellent performance across these diverse metrics indicates that our hybrid approach successfully addresses key limitations of existing single-model frameworks, providing a more robust and versatile solution for cross-view image synthesis tasks. The quantitative results presented in this section are compiled from multiple sources to ensure a fair and comprehensive comparison. Specifically, Pix2pix, X-Fork, and X-Seq results are obtained from~\cite{toker2021coming}, with their corresponding FID scores sourced from~\cite{wu2022cross_prog}. SelectionGAN and PAGAN results are taken from~\cite{wu2022cross_prog}. PGSL and SSVS metrics are sourced from~\cite{toker2021coming}, while SSVG results are from~\cite{shi2022geometry}. Cross-MLP results are obtained from~\cite{ren2021cascaded}, LGGAN from~\cite{Tang_2020_CVPR}, and PanoGAN from~\cite{wu2022cross}. Finally, CrossViewDiff and Sat2Density results are sourced from~\cite{li2024crossviewdiff}. This comprehensive compilation ensures that all comparisons are based on standardized evaluation protocols and consistent experimental conditions.

\begin{figure*}[tbp]
    \centering
     \resizebox{13cm}{!}{
   \hspace*{-3cm} 
    \begin{tabular}{ccc}
    \vspace{0.5cm}

    & (a) & (b) \\
       \multirow{-4}{*}{Satellite}& 
        \includegraphics[width=0.15\textwidth]{images/Fig8.jpg} &  
        \includegraphics[width=0.15\textwidth]{images/Fig18.jpg}

         \\ 
         
        \multirow{-4}{*}{PanoGAN\cite{wu2022cross}} &
        \includegraphics[width=10cm,height=2.5cm]{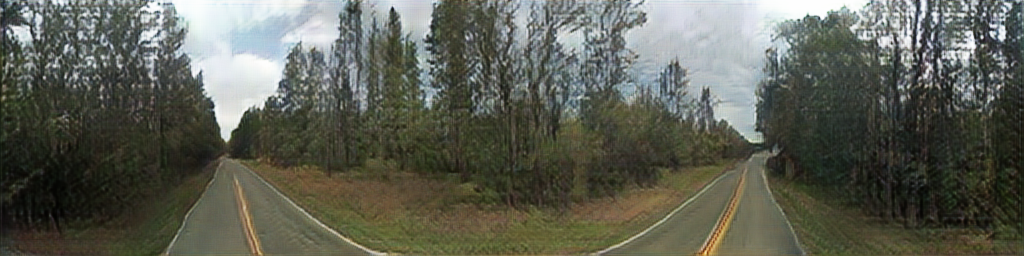} &
         \includegraphics[width=10cm,height=2.5cm]{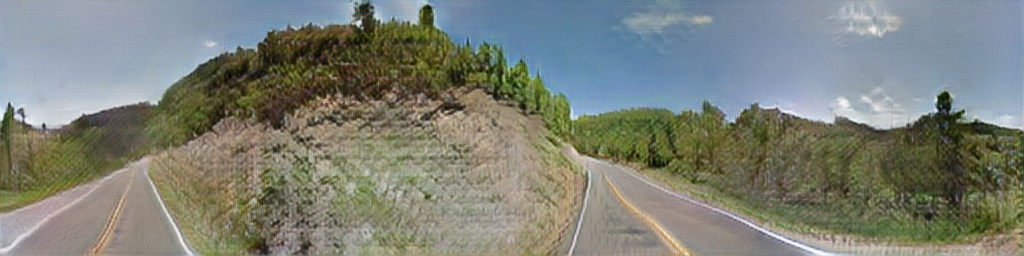}\\

              \multirow{-4}{*}{Sat2Density\cite{qian2023sat2density}} &
        \includegraphics[width=10cm,height=2.5cm]{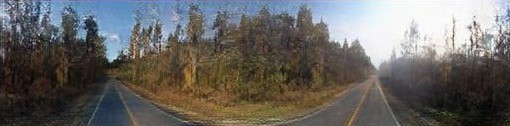} &
         \includegraphics[width=10cm,height=2.5cm]{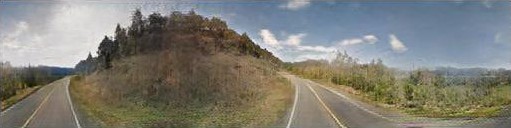}

         \\ 
         \multirow{-4}{*}{SSVS\cite{toker2021coming}} &  \includegraphics[width=10cm,height=2.5cm]{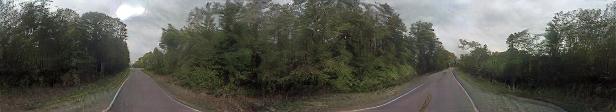} &
         \includegraphics[width=10cm,height=2.5cm]{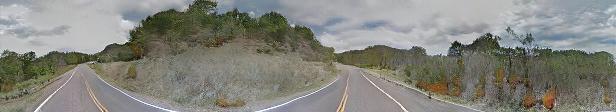}
         \\ 
          \multirow{-4}{*}{Ours} & \includegraphics[width=10cm,height=2.5cm]{images/Fig11.jpg} &
          \includegraphics[width=10cm,height=2.5cm]{images/Fig21.jpg}
         \\
         
            \multirow{-4}{*}{Ground Truth} & \includegraphics[width=10cm,height=2.5cm]{images/Fig12.jpg} &
            \includegraphics[width=10cm,height=2.5cm]{images/Fig22.jpg}

    \end{tabular}}
    \caption{Qualitative comparison with existing methods.}
    \label{fig:qualititave_com_pano_6}
    \end{figure*}

\begin{figure*}[tbp]
    \centering
     \resizebox{13cm}{!}{
   \hspace*{-3cm} 
    \begin{tabular}{ccc}
    \vspace{0.5cm}

    & (a) & (b) \\
       
       \multirow{-4}{*}{Satellite}& 
        \includegraphics[width=0.15\textwidth]{images/Fig28.jpg} &  \includegraphics[width=0.15\textwidth]{images/Fig33.jpg}

         \\ 
         
        \multirow{-4}{*}{PanoGAN\cite{wu2022cross}} &
        \includegraphics[width=10cm,height=2.5cm]{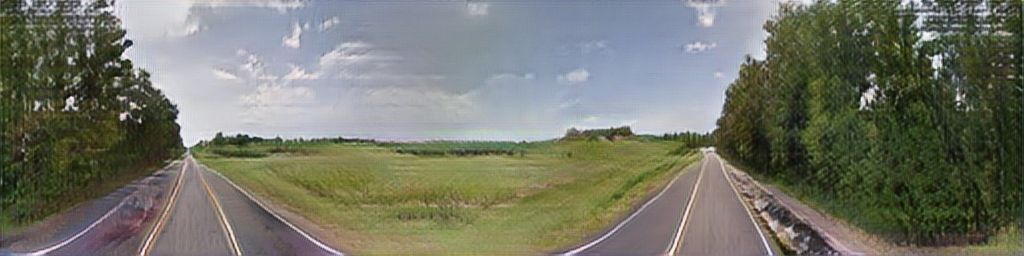} &
         \includegraphics[width=10cm,height=2.5cm]{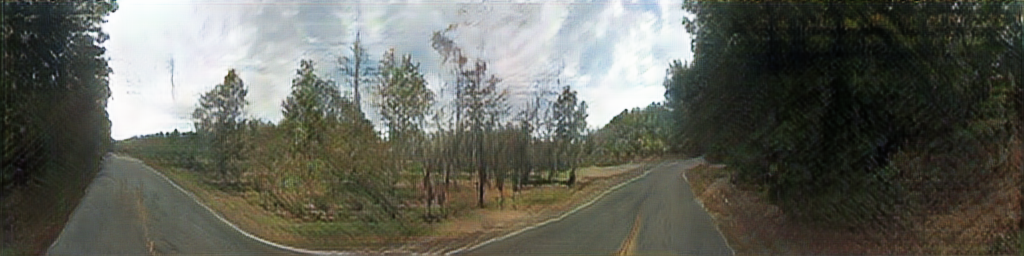}\\

                      \multirow{-4}{*}{Sat2Density\cite{qian2023sat2density}} &
        \includegraphics[width=10cm,height=2.5cm]{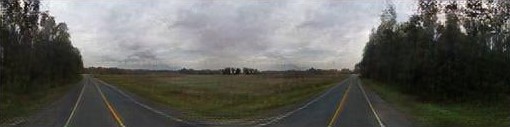} &
         \includegraphics[width=10cm,height=2.5cm]{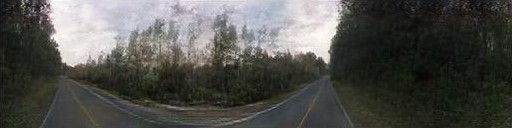} 
        
         \\ 
         \multirow{-4}{*}{SSVS\cite{toker2021coming}} &  \includegraphics[width=10cm,height=2.5cm]{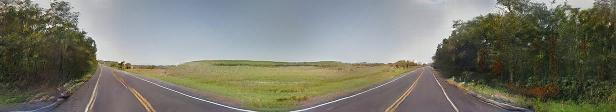} &
         \includegraphics[width=10cm,height=2.5cm]{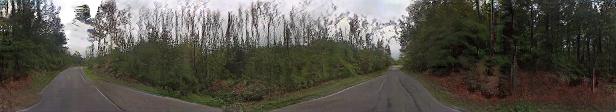}
         \\ 
          \multirow{-4}{*}{Ours} & \includegraphics[width=10cm,height=2.5cm]{images/Fig31.jpg} &
          \includegraphics[width=10cm,height=2.5cm]{images/Fig36.jpg}
         \\
         
            \multirow{-4}{*}{Ground Truth} & \includegraphics[width=10cm,height=2.5cm]{images/Fig32.jpg} &
            \includegraphics[width=10cm,height=2.5cm]{images/Fig37.jpg}

    \end{tabular}}
    \caption{Qualitative comparison with existing methods.}
    \label{fig:qualititave_com_pano_7}
    \end{figure*}

\noindent\textbf{Qualitative Comparison.} Figures \ref{fig:qualti-comparision-square}, \ref{fig:qualititave_com_pano_6}, and \ref{fig:qualititave_com_pano_7} present a qualitative comparison of our framework against existing methods on the test set of the CVUSA dataset. We compare our proposed framework with five methods, which include both square and panorama street-view images: LGGAN \cite{Tang_2020_CVPR}, SelectionGAN \cite{tang2019multi}, PanoGAN \cite{wu2022cross}, Sat2density \cite{qian2023sat2density}, and SSVS \cite{toker2021coming}. Overall, our model generates geometrically consistent street-view images with fewer artifacts. Also,  it captures finer local details, such as clouds and street lines, compared to existing methods, bringing the generated images visually closer to the ground truth. Compared to SSVS and Sat2density, our framework generates better images in terms of quality level with finer details in several cases. While PanoGAN generates comparable quality in some cases, it introduces noticeable artifacts in some images, as shown in \ref{fig:qualititave_com_pano_7} (a). Additionally, our framework captures cloud patterns more effectively, as illustrated in Figures \ref{fig:qualititave_com_pano_6} (b), and \ref{fig:qualititave_com_pano_7} (a), and generates sharper street lines, as seen in Figures \ref{fig:qualititave_com_pano_6} (a), and \ref{fig:qualititave_com_pano_6} (b). SelectionGAN and LGGAN generate low-quality images with visible artifacts. While some street lines are present in some cases, they are not well-defined. The comparison shows that our framework maintains both image quality and captures fine details, while other methods tend to either generate high-quality images with some local details being ignored or capture details with degraded overall quality. A distinguishing feature of our proposed framework is its capability to include parts of fence structures during the generation process, which are being missed in the results generated by other methods, as illustrated in Figure \ref{fig:qualititave_com_pano_7} (b).

\begin{table*}[tbp]
\centering
\caption{Quantitative comparison on CVUSA dataset.}
\label{tab:Quantitative_comp}
\renewcommand{\arraystretch}{1.5}
\setlength{\tabcolsep}{8pt}
\begin{tabular}{lcccc}
\hline \hline
\multirow{2}{*}{Method} & \multicolumn{4}{c}{Evaluation Metrics} \\
\cline{2-5}
& SSIM $\uparrow$ & PSNR $\uparrow$ & FID $\downarrow$ & LPIPS $\downarrow$ \\ 
\hline 
Pix2Pix \cite{isola2017} & 0.392 & 11.671 & 258.2 & 0.595 \\
X-Fork \cite{regmi2018} & 0.435 & 13.064 & 258.2 & 0.609 \\
X-Seq \cite{regmi2018} & 0.423 & 12.820 & 274.16 & 0.590 \\
SelectionGAN \cite{tang2019multi} & 0.4172 & 19.3237 & 247.76 & - \\
PAGAN \cite{wu2022cross_prog} & 0.4315 & 19.4518 & 99.54 & - \\
SSVG \cite{shi2022geometry} & 0.3408 & 13.77 & - & - \\
PGSL \cite{zhai2017} & 0.414 & 11.502 & - & - \\
SSVS \cite{toker2021coming} & 0.447 & 13.895 & - & 0.474 \\
Cross-MLP \cite{ren2021cascaded} & 0.5251 & 23.1532 & - & - \\
PanoGan \cite{wu2022cross} & 0.4437 & 20.9467 & - & - \\
LGGAN \cite{Tang_2020_CVPR} & 0.5238 & 22.5766 & - & - \\
CrossViewDiff \cite{li2024crossviewdiff} & 0.371 & 12.000 & 23.67 & - \\
Sat2Density \cite{qian2023sat2density} & 0.339 & 14.229 & 41.43 & - \\
\hline
\textbf{Ours} & \textbf{0.3464} & \textbf{13.41} & \textbf{40.32} & \textbf{0.6305} \\
\hline \hline
\end{tabular}

\end{table*}


\subsection{Ablation Study}
In this section, we conduct an ablation study on the CVUSA dataset to assess the impact of various input prompts on the Stable Diffusion component. By performing that, we can understand how prompt design affects the overall performance of our framework. \newline 
\subsubsection{Integrating LVLM with Stable Diffusion model}

\label{ab:integrating}
We investigated three approaches for providing input prompts to the Stable Diffusion model. 
\begin{itemize}
    \item \textbf{No LVLM (fixed prompt):} In this configuration, the "street view" caption is utilized as a fixed prompt for all images. 
    \item \textbf{General-purpose LVLM:} In this setup, we applied a general-purpose BLIP-based LVLM, called FuseCap \cite{rotstein2024fusecap} for generating image captions that were used as input prompts for the Stable Diffusion model.
    
    \item \textbf{Domain-specific LVLM:} In this setup, we employed RS-LLaVa, which is an LLaVA-based LVLM, adopted from \cite{bazi2024rs}. It has been fine-tuned to generate captions for satellite images. These captions were then used as input prompts for our Stable Diffusion model.
    
\end{itemize}

\begin{figure*}[tbp]
\hspace{2cm}
\resizebox{11cm}{!}{
    \begin{tabular}{p{7cm}c}

\multirow{-6}{*}{ a) Satellite Image } & \includegraphics[width=3cm]{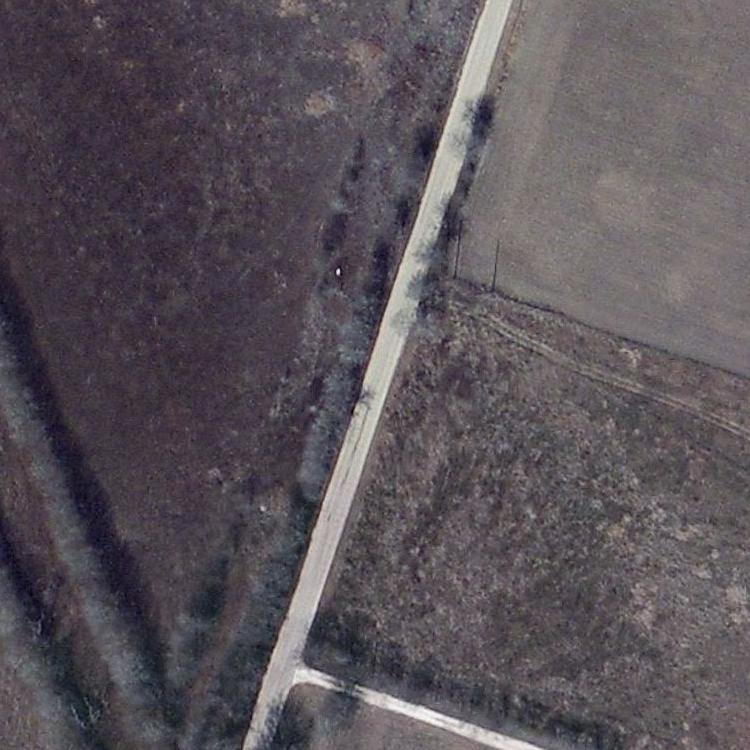} \\
    
   \multirow{-6}{*}{ b) Ground Truth} & \includegraphics[width=12cm,height=3cm]{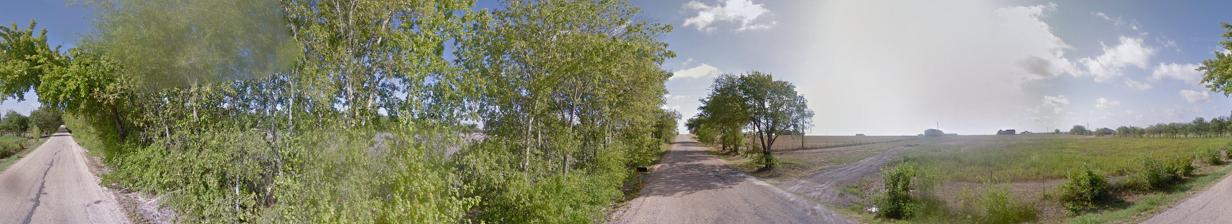} \\
    
\multirow{-6}{*}{c) Input Prompt: "Street view"} & \includegraphics[width=12cm,height=3cm]{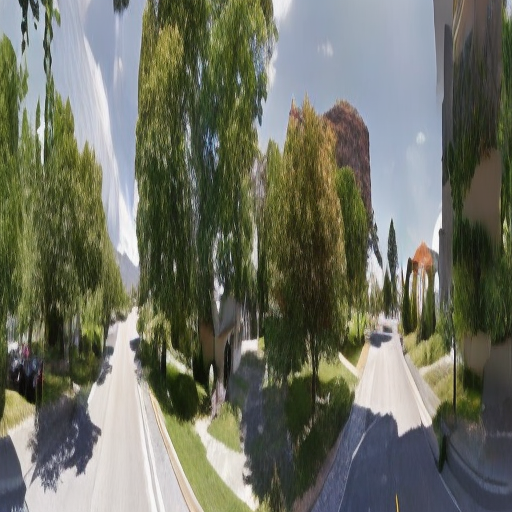} \\
\vspace{-2cm}
   \fontsize{10}{12}\selectfont  d) Input Prompt (Fuse Cap) :  " a parking lot featuring a row of parked cars, a building with a clock tower, and a green tree in the background"
     
& \includegraphics[width=12cm,height=3cm]{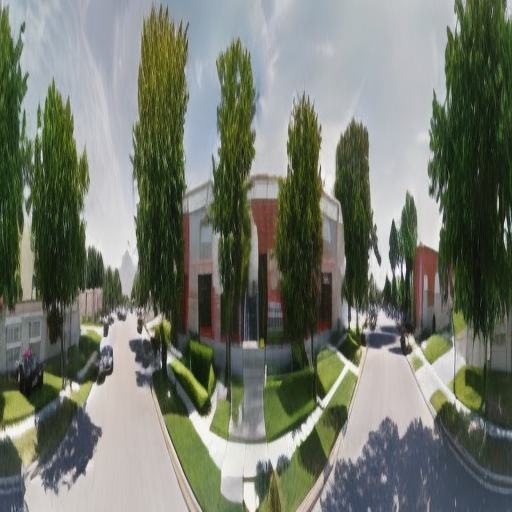} \\
\vspace{-2.75cm}
       \fontsize{10}{12}\selectfont e) Input Prompt (RS-LLaVA): " The image depicts a scene with two roads intersecting, surrounded by bare land and a few trees. The main objects are the roads, trees, and the open land. The roads are located at the center of the image, while the trees are scattered around the edges. The bare land occupies the majority of the scene, with the roads and trees being smaller elements within it."
      & \includegraphics[width=12cm,height=3cm]{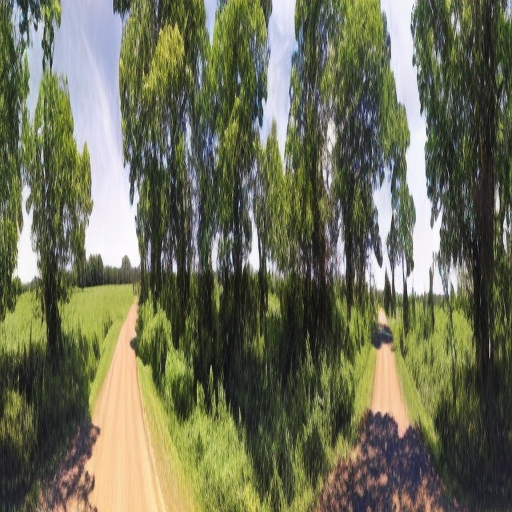} \\
    
\end{tabular}}
    \caption{Qualitative comparison of various input prompt approaches on the CVUSA dataset }
    \label{fig:ab_1}
\end{figure*}

\begin{figure*} [tbp]
\hspace{2cm}
\resizebox{11cm}{!}{
    \begin{tabular}{p{7cm}c}

   \multirow{-6}{*}{ a) Satellite Image } & \includegraphics[width=3cm]{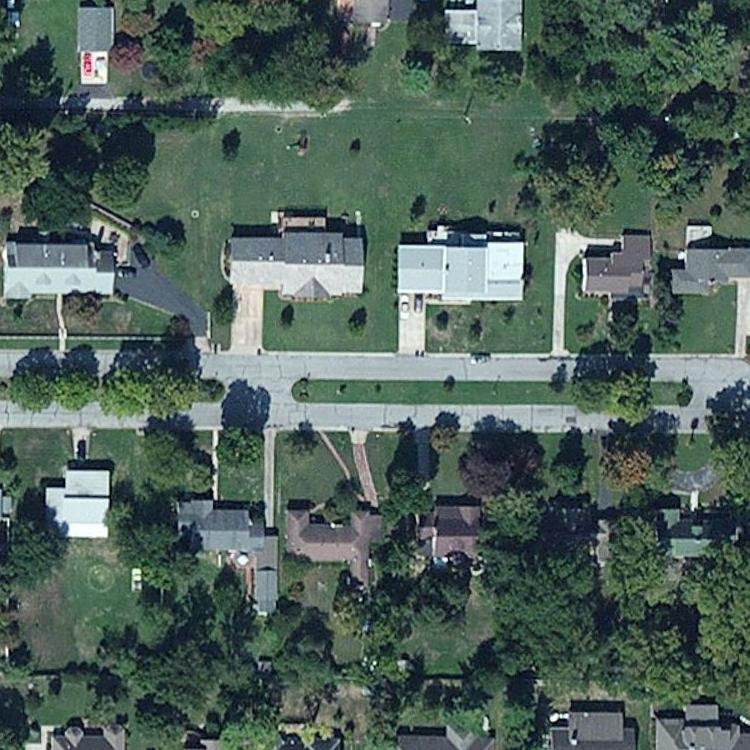} \\
    
   \multirow{-6}{*}{b) Ground Truth }& \includegraphics[width=12cm,height=3cm]{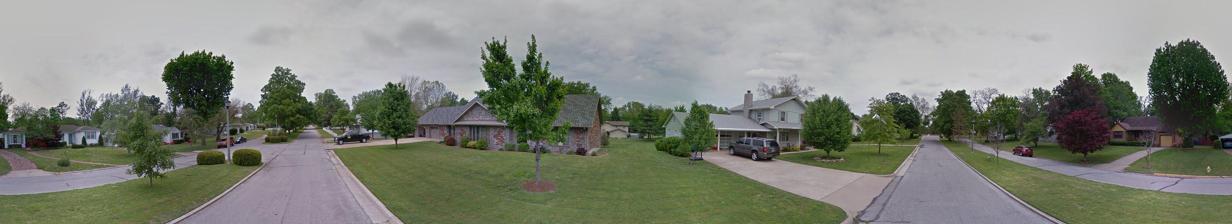} \\
    
    \multirow{-6}{*}{c) Input Prompt: "Street view"} & \includegraphics[width=12cm,height=3cm]{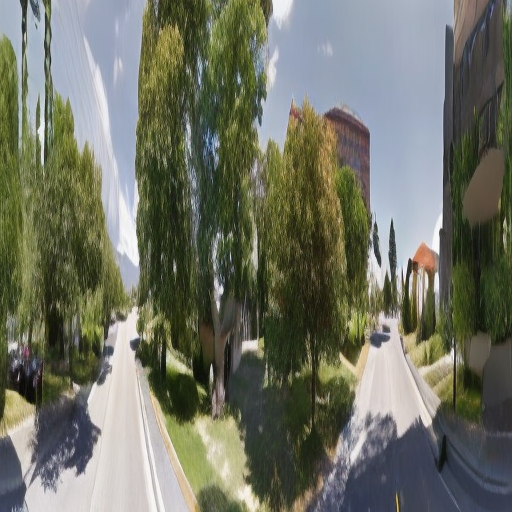} \\
    \vspace{-2.5cm}
    \fontsize{10}{12}\selectfont d) Input Prompt (Fuse Cap): "a white building stands tall amidst a lush green landscape, with a road leading up to it and two trees - one large and one small - adding to the natural beauty of the"
 & \includegraphics[width=12cm,height=3cm]{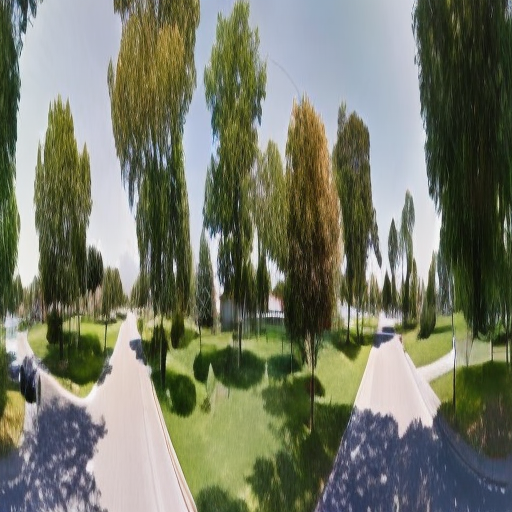} \\
 \vspace{-2.75cm}
      \fontsize{10}{12}\selectfont e) Input Prompt (RS-LLaVA): 
      "The image features a residential area with numerous buildings, green trees, and a road running through it. The buildings are spread across the landscape, with some clustered together and others standing alone. The trees are interspersed among the buildings, providing a natural contrast to the urban setting. The road serves as a connector between the various parts of the neighborhood, allowing for easy access and movement within the area."

      & \includegraphics[width=12cm,height=3cm]{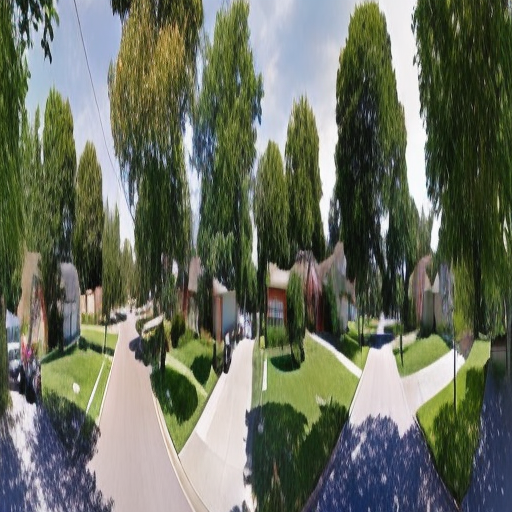} \\
    
\end{tabular}}
    \caption{Qualitative comparison on variant input prompt approaches on CVUSA dataset }
    \label{fig:ab_2}
\end{figure*}

\begin{figure*}[tbp]
\hspace{2cm}
\resizebox{11cm}{!}{

    \begin{tabular}{p{7cm}c}

   \multirow{-6}{*}{ a) Satellite Image } & \includegraphics[width=3cm]{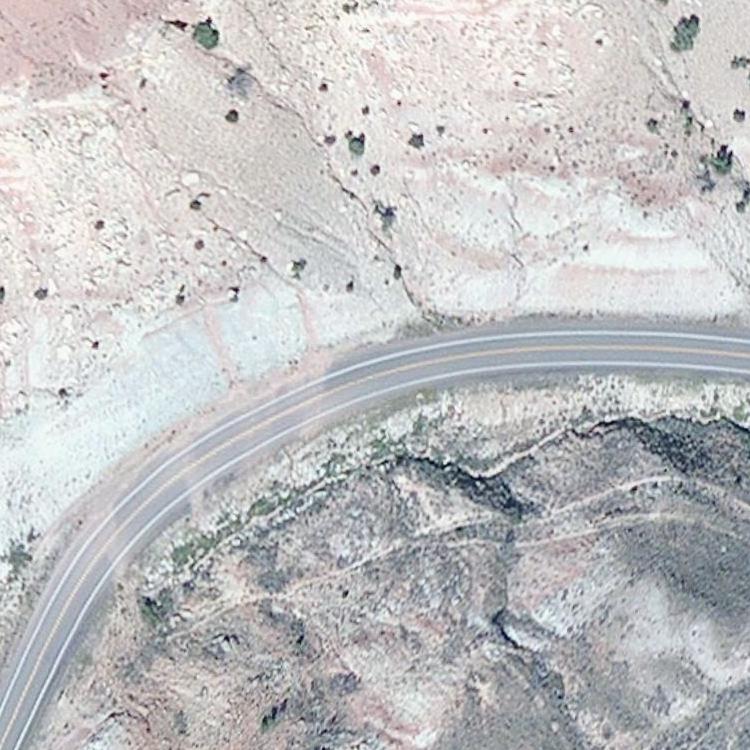} \\
    
   \multirow{-6}{*}{ b) Ground Truth} & \includegraphics[width=12cm,height=3cm]{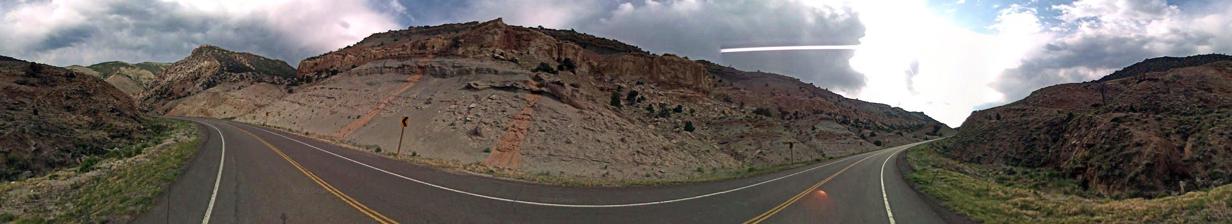} \\
    
    \multirow{-6}{*} {c) Input Prompt: "Street view"} & \includegraphics[width=12cm,height=3cm]{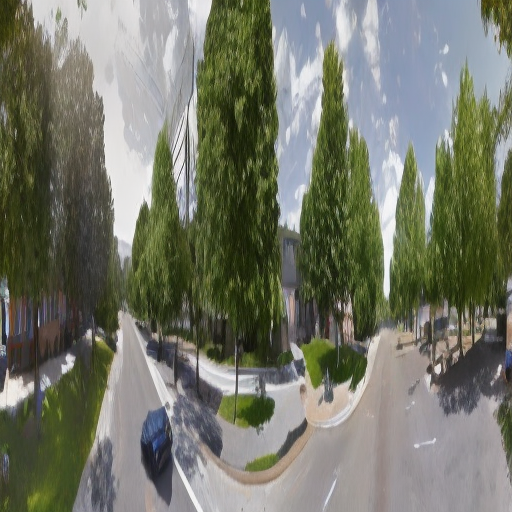} \\
    \vspace{-2cm}
      \fontsize{10}{12}\selectfont d) Input Prompt (Fuse Cap): "a winding road cuts through a mountainous landscape, with a clear blue sky overhead and a distant mountain range visible in the distance."

& \includegraphics[width=12cm,height=3cm]{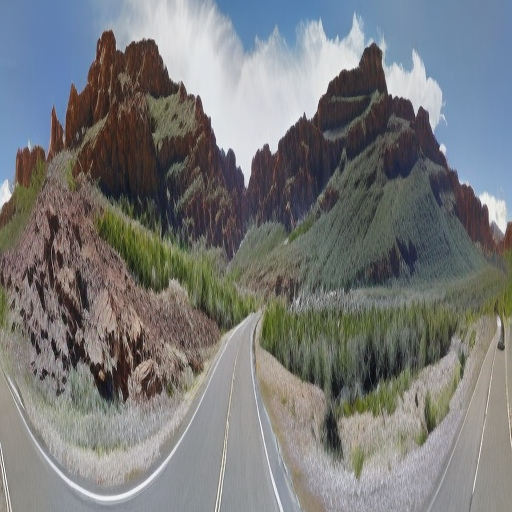} \\
\vspace{-2.75cm}
       \fontsize{10}{12}\selectfont e) Input Prompt (RS-LLaVA): "The image depicts a road winding through a desert landscape, with a few trees scattered along its path. The main objects in the scene are the road, the desert, and the trees. The road is located in the center of the image, while the desert surrounds it on both sides. The trees are dispersed throughout the desert landscape and can be seen along the road's path."

      & \includegraphics[width=12cm,height=3cm]{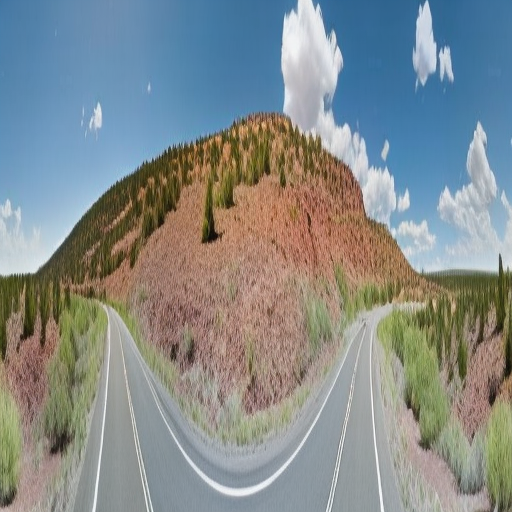} \\
    
\end{tabular}}
    \caption{Qualitative comparison on variant input prompt approaches on CVUSA dataset }
    \label{fig:ab_3}
\end{figure*}

By considering these three approaches as a source of input prompts for the Stable Diffusion model, we observe that using a "street view" fixed prompt results in the generation of general cityscape images that contain elements such as trees, buildings, and cars. However, these generated images are not geometrically consistent with the given satellite images. In addition, we can see that the generated images include details not present in the ground-truth street-view image, suggesting that the model generates additional creative objects beyond the given data. Figures \ref{fig:ab_1} (c), \ref{fig:ab_2} (c), and \ref{fig:ab_3} (c). show a sample of generated street-view images using the "street view" prompt.

For the general-purpose LVLM (FuseCap), we note that for some images, this LVLM model generates captions that do not accurately describe the provided satellite image. For example, in Figure \ref{fig:ab_1} (d), the incorrect caption mentions the parking lot, cars, and clock tower, which does not exist in the given satellite image, resulting in an inconsistent street-view image. Additionally, the general captions generated by the FuseCap LVLM result in street-view images with considerable differences in the road layouts, as shown in Figure \ref{fig:ab_3} (d), and missing buildings, as seen in Figure \ref{fig:ab_2} (d). 

For the domain-specific LVLM (RS-LLaVA), we observe that this LVLM is capable of generating more expressive captions for satellite images by describing more details and providing positional information about the objects. This results in a better overall representation and more geometrically consistent street-view images. As an example, in Figure \ref{fig:ab_1} (e), the caption specifies that "trees are scattered" and "open land" are present in the satellite image, and it identifies the main objects in the image. In addition, the input prompt for Figure \ref{fig:ab_2} (e) highlights several details like "numerous buildings," "buildings are spread across the landscape," and " trees are interspersed among the buildings." Although the resulting street-view image is not entirely geometrically consistent, the presence of houses, trees, and roads shapes in the generated image creates a more residential environment, bringing it closer to the ground compared to the street-view images generated using a fixed prompt and general-purpose LVLM, as seen in Figures \ref{fig:ab_2} (c) and \ref{fig:ab_2} (d), respectively. Moreover, Figure  \ref{fig:ab_3} (e) demonstrates a significant improvement in the generated street view for the satellite image in Figure \ref{fig:ab_3} (a). The RS-LLaVA model effectively generates a caption describing the satellite image, including some details such as "desert landscape," "few trees scattered along its path," and "the road is located in the center." It also identifies the main objects in the satellite images, leading to the generating of a more consistent and accurate street view in Figure  \ref{fig:ab_3} (e)  compared to the images in Figures \ref{fig:ab_3} (c) and \ref{fig:ab_3} (d). 



\section{Conclusion}\label{conclusion}
Street-view imagery serves as a valuable resource across multiple domains, providing important insights for urban analytics and decision-making processes. Despite existing methods in this field, a gap remains in leveraging state-of-the-art deep learning techniques for cross-view synthesis. To address this limitation, we developed a novel hybrid framework that combines the strengths of diffusion-based models and conditional generative adversarial networks for satellite-to-street-view synthesis. Through a comprehensive quantitative and qualitative evaluation on the CVUSA dataset, our proposed framework demonstrates a good capability in generating geometrically consistent street-view images that closely approximate real-world scenarios, successfully capturing fine-grained local details, including clouds, street markings, and architectural elements such as fence structures. The framework's ability to generate realistic and consistent street-view images from satellite imagery opens new opportunities for improving geo-localization tasks by creating consistent satellite-street-view image pairs for accurate location retrieval. Additionally, the synthetic street-view imagery can serve as valuable training data for deep learning models in autonomous driving, robotics, and urban planning systems. The challenging nature of cross-view synthesis, due to limited appearance overlap between satellite and street-view domains, continues to present opportunities for further improvements and remains an active area of research. As future work, we plan to integrate recently developed advanced generative models, including next-generation diffusion architectures such as FLUX~\cite{flux2024} and improved stable diffusion variants, to replace conditional GAN components. Additionally, we intend to explore advanced sampling methods and enhanced denoising processes within the diffusion model framework. Moreover, we plan to expand dataset diversity by incorporating street-view images from broader geographical regions and various weather conditions, including rain, snow, and different lighting scenarios, to improve the framework's robustness and real-world applicability.
\IEEEpeerreviewmaketitle

\bibliographystyle{IEEEtran}
\bibliography{main}
\end{document}